\batchmode
\makeatletter
\def\input@path{{"C:/Trabajo laptop/Mis articulos/Finished/TGOSPA q-metric/Accepted/"}}
\makeatother
\documentclass[onecolumn,english,twocolumn]{IEEEtran}
\usepackage[T1]{fontenc}
\usepackage[latin9]{inputenc}
\usepackage{color}
\usepackage{array}
\usepackage{multirow}
\usepackage{varwidth}
\usepackage{amsmath}
\usepackage{amsthm}
\usepackage{amssymb}
\usepackage{graphicx}

\makeatletter

\providecommand{\tabularnewline}{\\}
\newenvironment{cellvarwidth}[1][t]
    {\begin{varwidth}[#1]{\linewidth}}
    {\@finalstrut\@arstrutbox\end{varwidth}}

\theoremstyle{plain}
\newtheorem{thm}{\protect\theoremname}
\theoremstyle{definition}
\newtheorem{defn}[thm]{\protect\definitionname}
\theoremstyle{plain}
\newtheorem{lem}[thm]{\protect\lemmaname}
\theoremstyle{plain}
\newtheorem{prop}[thm]{\protect\propositionname}
\theoremstyle{definition}
\newtheorem{example}[thm]{\protect\examplename}

\usepackage{cite} 
\usepackage[margin=8pt,font=footnotesize]{caption}
\usepackage{algorithm}
\usepackage{algpseudocode}

\usepackage{amsmath}  
\allowdisplaybreaks

\makeatother

\usepackage{babel}
\providecommand{\definitionname}{Definition}
\providecommand{\examplename}{Example}
\providecommand{\lemmaname}{Lemma}
\providecommand{\propositionname}{Proposition}
\providecommand{\theoremname}{Theorem}

\begin{document}
\title{GOSPA and T-GOSPA quasi-metrics for evaluation of multi-object tracking
algorithms}
\author{Ángel F. García-Fernández, Jinhao Gu, Lennart Svensson, Yuxuan Xia,
Jan Krej\v{c}í, Oliver Kost, Ond\v{r}ej Straka\thanks{A. F. Garc\'ia-Fern\'andez is with the Information Processing and Telecommunications Center, ETSI de Telecomunicaci\'on, Universidad Polit\'ecnica de Madrid, 28040 Madrid, Spain (email: \mbox{angel.garcia.fernandez@upm.es}).} 
\thanks{J. Gu is with the Department of Electrical Engineering and Electronics, University of Liverpool, Liverpool L69 3GJ (email: jinhgu@liverpool.ac.uk).}
\thanks{L. Svensson is with the Department of Electrical Engineering, Chalmers University of Technology, SE-412 96 Gothenburg, Sweden (email: lennart.svensson@chalmers.se).} 
\thanks{Y. Xia is with the Department of Automation and Intelligent Sensing, Shanghai Jiao Tong University, Shanghai, China (e-mail:
\mbox{yuxuan.xia@sjtu.edu.cn}).} 
\thanks{J. Krej\v{c}\'i, O. Kost, O. Straka are with the Department of Cybernetics, University of West Bohemia, Pilsen, Czech Republic (e-mails: \{jkrejci, kost, straka30\}@kky.zcu.cz).}
\thanks{This work was supported by the Spanish Ministry of Science, Innovation and Universities under the project PID2024-158149OB-C21.}}

\maketitle

\begin{abstract}
This paper introduces two quasi-metrics for performance assessment
of multi-object tracking (MOT) algorithms. One quasi-metric is an
extension of the generalised optimal subpattern assignment (GOSPA)
metric and measures the discrepancy between sets of objects. The other
quasi-metric is an extension of the trajectory GOSPA (T-GOSPA) metric
and measures the discrepancy between sets of trajectories. Similar
to the GOSPA-based metrics, these quasi-metrics include costs for
localisation error for properly detected objects, the number of false
objects and the number of missed objects. The T-GOSPA quasi-metric
also includes a track switching cost. Differently from the GOSPA and
T-GOSPA metrics, the proposed quasi-metrics have the flexibility of
penalising missed and false objects with different costs, and the
localisation costs are not required to be symmetric. We also explain
how to obtain similarity score functions based on these quasi-metrics.
The performance of several Bayesian MOT algorithms is assessed with
the T-GOSPA quasi-metric via simulations. 

\end{abstract}


\section{Introduction}

Multi-object tracking (MOT) consists of estimating the trajectories
of a variable and an unknown number of objects using sensor measurements.
MOT has a wide variety of applications such as traffic monitoring
\cite{Jimenez-Bravo22}, underwater surveillance \cite{Braca14},
and space object cataloging \cite{Delande19}. This paper considers
the important topic of performance evaluation of MOT algorithms \cite{Blackman_book99,Smith05}.
That is, given a ground truth set of trajectories, and the estimated
sets of trajectories provided by several algorithms, the best performing
algorithm is the one whose estimate is the most similar to the ground
truth. Therefore, to assess MOT algorithm performance, it is necessary
to establish a definition of error or distance between two sets of
trajectories. 

In mathematics, the notion of error or distance can be defined via
metrics\footnote{The terms distance and metric are sometimes used interchangeably.
In this paper, a distance refers to a function that may not fulfill
all the mathematical properties required of a metric.}. Metrics are non-negative functions that meet three properties: identity,
symmetry, and triangle inequality \cite{Apostol_book74}. These properties
ensure that the notion of distance is intuitive and they are the foundation
of important properties for mathematical analysis, such as the concept
of metric spaces. The triangle inequality guarantees that distances
between points follow a consistent and logical pattern. That is, it
prevents cases where the direct distance between two endpoints is
greater than the sum of the distances between any intermediate points.
We proceed to review the literature on performance evaluation for
MOT.

Multi-object filtering is a sub-problem of MOT in which the goal is
to estimate the current set of objects, rather than the set of trajectories.
A widely-used metric for sets of objects is the optimal sub-pattern
assignment (OSPA) metric \cite{Schuhmacher08_b,Schuhmacher08}. Other
metrics for sets of objects are the cardinalised optimal linear assignment
(COLA) metric \cite{Barrios17,Barrios23} (which is equal to the unnormalised
OSPA (UOSPA) metric \cite[Eq. (44)]{Williams15} divided by its cardinality
mismatch parameter $c$), the Hausdorff and Wasserstein metrics \cite{Hoffman04},
and the complete OSPA metric \cite{Vu20}. However, none of the above
mentioned metrics penalise all the factors that are considered of
interest in MOT: localisation errors for detected objects, the number
of missed objects and the number of false objects, in the sense that
an increase in any of these factors implies an increase in the metric
value. For instance, in some situations, we can add false objects
to the estimated set of objects, but the values of the above metrics
remain unchanged \cite[Example 2]{Rahmathullah17}. These factors
can be penalised with the generalised optimal sub-pattern assignment
(GOSPA) metric \cite{Rahmathullah17}. 

In MOT, apart from penalising the above factors, it is relevant to
penalise track switches since the estimated trajectories of different
objects may be swapped at some point in time. In computer vision,
there are several methods to assess the accuracy of the estimated
trajectories, including the number of track switches, which can be
lowered by object re-identification modules \cite{Suljagic22,Bayraktar22,Bayraktar25}.
One example is the multiple object tracking accuracy (MOTA) score
\cite{Bernardin08}, and its variations \cite{Caesar20}. MOTA is
based on obtaining a heuristic matching between the ground truth states
and the estimated object states at each time step. With this matching,
the MOTA score is defined based on the number of track switches, missed
and false objects, but not on the localisation errors for matched
objects \cite{Bernardin08}. The MOTA score indicates similarity,
takes values in the interval $\left(-\infty,1\right]$, and is non-symmetric.
As MOTA does not consider localisation errors, these are usually measured
in this setting by calculating a companion index, the multiple object
tracking precision (MOTP). The MOTP was originally defined as an error
\cite{Bernardin08}, but it can also be defined as a similarity with
a minor modification \cite{Luiten20}. 

Another similarity score for MOT used in computer vision, taking values
in $\left[0,1\right]$, is the higher order tracking accuracy (HOTA)
score \cite{Luiten20}. The HOTA score for a given similarity threshold
$\alpha$ ($\mathrm{HOTA}_{\alpha}$) is based on solving an external
assignment problem (e.g., an assignment problem with another cost
function) at each time step. The solutions of the external assignment
problems are then used to calculate the $\mathrm{HOTA}_{\alpha}$
score, which depends on the number of missed objects, false objects
and track switches, but it does not depend on localisation errors
for properly detected objects. The overall HOTA is then obtained by
averaging $\mathrm{HOTA}_{\alpha}$ over multiple similarity thresholds,
to indirectly account for localisation errors. However, none of MOTA,
MOTP or HOTA are mathematical metrics, as they do not meet the identity
and triangle inequality properties.  

To define a metric for sets of trajectories, one solution is to first
determine a base distance for trajectories and then apply the OSPA
metric. This procedure results in the OSPA$^{(2)}$ metric \cite{Beard20}.
It is also direct to prove that an unnormalised version of the OSPA$^{(2)}$
metric, referred to as UOSPA$^{(2)}$, is also a metric. OSPA$^{(2)}$
and UOSPA$^{(2)}$ associate estimated trajectories to ground truth
trajectories, but the association remains fixed across time. Therefore,
OSPA$^{(2)}$ and UOSPA$^{(2)}$ do not include penalties for track
switches, and do not penalise the number of missed and false objects,
either \cite{Angel21_f}. 

Bento's metrics are mathematically consistent metrics for sets of
trajectories and penalise track switches, though they require the
introduction of $\ast$-trajectories to ensure the two sets have the
same number of elements \cite{Bento_draft16}. The extension of the
GOSPA metric to sets of trajectories, termed trajectory GOSPA (T-GOSPA)
metric, was presented in \cite{Angel20_d} to penalise localisation
errors, the number of missed and false objects, and the number of
track switches. T-GOSPA works by assigning at each time step trajectories
in both sets, while allowing for the possibility of leaving trajectories
unassigned. T-GOSPA was extended to have time-weighted costs in \cite{Angel21_f}.
Both T-GOSPA and Bento's metrics have linear programming (LP) relaxation
variants, which are also metrics and are faster to compute. Approximate,
fast implementations of the T-GOSPA metric for large tracking scenarios
based on unbalanced multimarginal optimal transport and graph structured
optimal transport have been recently proposed in \cite{Wernholm25,Warnsater25}.

A property of the GOSPA and T-GOSPA metrics is that missed and false
objects are penalised with the same cost, to ensure the symmetry property.
The previous distances and similarity scores also share this property.
However, there are some applications in which it is more suitable
to have different costs for missed and false objects. For instance,
in classic radar detector design, one typically sets a very low probability
of false alarm and then maximises the probability of detection using
a Neyman-Pearson test \cite[Chap. 10]{Skolnik_book01}. With this
design, the cost of a false object is higher than the cost of a missed
object. In other applications, it is crucial not to miss any objects,
for instance, in self-driving vehicles, unmanned surface vehicles
or satellite collision avoidance systems. Hence, it is desirable to
assess algorithms with a higher penalty for missed objects than for
false objects. 

To achieve this flexibility in the costs for missed and false objects,
it is necessary to develop distance functions that are non-symmetric.
Distances that meet the metric properties except the symmetry property
are called quasi-metrics (q-metrics) \cite{Wilson31,Schroeder06,Bento19}.
Q-metrics have been used in several applications: similarity search
in protein datasets \cite{Stojmirovic_thesis05}, reinforcement learning
\cite{Wang23}, and q-metric learning \cite{Wang22}. A q-metric for
MOT with continuous trajectories has been proposed in \cite{Li26},
the Star-ID metric. The Star-ID metric penalises trajectory segments
not assigned to other trajectories as well as full unassigned trajectories,
with costs depending on the segment/trajectory lengths. However, the
Star-ID metric has the constraint of requiring fixed assignments of
trajectories across time, and therefore it does not include costs
for track switches. In addition, it is not applicable to the standard
MOT problems that estimate trajectories in discrete time. This paper
proposes q-metrics for multi-object filtering and MOT that penalise
the aspects of interest in MOT: localisation errors, number of false
objects, number of missed objects, and track switches. The proposed
GOSPA and T-GOSPA q-metrics are designed based on the GOSPA and T-GOSPA
metrics and use a base distance that is a q-metric and incorporate
an additional parameter $\rho\in\left(0,1\right)$ that controls the
penalties for false and missed objects. In addition, motivated by
the use of similarity score functions, taking values in $\left[0,1\right]$,
to measure performance in MOT in computer vision, we define and derive
score functions based on the proposed q-metrics to measure similarity
between sets of objects and trajectories.

The contributions of this paper are:
\begin{itemize}
\item Development of the GOSPA q-metric\footnote{Matlab and Python implementations of the GOSPA and T-GOSPA q-metrics
are available at https://github.com/Agarciafernandez/MTT and https://github.com/Agarciafernandez/T-GOSPA-metric-python,
respectively.} for sets of objects for performance evaluation of multi-object filters
with different costs for missed and false objects, including the proofs
of the q-metric properties. 
\item Development of the T-GOSPA q-metric for sets of trajectories for performance
evaluation of MOT algorithms with different costs for missed and false
objects, including the proofs of the q-metric properties.
\item Derivation of three properties of the GOSPA and T-GOSPA q-metrics
related to the q-metric parameter $\rho$. 
\item Definition and derivation of q-metric-based score functions, based
on GOSPA and T-GOSPA, to measure the similarity between sets of objects
and trajectories.
\item Extensions of the GOSPA and T-GOSPA q-metrics, and their associated
score functions, to random finite sets (RFSs) \cite{Mahler_book14}.
\end{itemize}
The paper also provides simulation results evaluating state-of-the-art
Bayesian MOT algorithms via the T-GOSPA q-metric. 

The organisation of the remainder of the paper is as follows. Section
\ref{sec:Background} presents the required background. Sections \ref{sec:GOSPA-quasi-metric}
and \ref{sec:Trajectory-GOSPA-quasi-metric} introduce the GOSPA and
T-GOSPA q-metrics, respectively. The derivation of q-metric-based
score functions is addressed in Section \ref{sec:Q-metric-based-scores}.
The extensions of the q-metrics to RFSs are provided in Section \ref{sec:Quasi-metrics-RFS}.
Simulation results are analysed via the T-GOSPA q-metric in Section
\ref{sec:Simulations}. Section \ref{sec:Conclusions} gives the conclusions.

\section{Background\label{sec:Background}}

This section introduces the variables we consider (Section \ref{subsec:Variables}),
the definitions of metrics and q-metrics (Section \ref{subsec:Metric-and-q-metrics}),
and the GOSPA metric (Section \ref{subsec:GOSPA-metric}).

\subsection{Variables\label{subsec:Variables}}

We denote a single object state as $x\in\mathbb{X}$, with $\mathbb{X}$
being the single-object space. The single-object space is typically
$\mathbb{X}=\mathbb{R}^{n_{x}}$, and contains information on the
object kinematics such as position and velocity, and it can also include
discrete variables such as object type. A set of objects is then represented
as $\mathbf{x}=\left\{ x_{1},...,x_{n_{\mathbf{x}}}\right\} $, and
its cardinality is $|\mathbf{x}|=n_{\mathbf{x}}$ \cite{Mahler_book14}. 

The trajectory of an object is denoted by $X=\left(\omega,x^{1:\nu}\right)$.
Here, $\omega$ is the trajectory start time step, $\nu$ is the trajectory
duration and $x^{1:\nu}=\left(x^{1},...,x^{\nu}\right)$ is the sequence
of object states of this trajectory \cite{Angel20_b}. We focus on
trajectories contained in a time window from time step 1 to time step
$T$, such that $1\leq\omega\leq T$ and $1\leq\nu\leq T-\omega+1$.
 A set of object trajectories is denoted by $\mathbf{X}=\left\{ X_{1},...,X_{n_{\mathbf{X}}}\right\} $,
with cardinality $|\mathbf{X}|=n_{\mathbf{X}}$. 

\subsection{Metric and q-metrics\label{subsec:Metric-and-q-metrics}}

A metric on a given space $\Upsilon$ is a non-negative function $d\left(\cdot,\cdot\right):\Upsilon\times\Upsilon\rightarrow\left[0,\infty\right)$
that meets the following properties for any $\mathbf{X},\mathbf{Y},\mathbf{Z}\in\Upsilon$
\cite{Apostol_book74}
\begin{itemize}
\item $d\left(\mathbf{X},\mathbf{Y}\right)=0$ if and only if $\mathbf{X}=\mathbf{Y}$
(identity),
\item $d\left(\mathbf{X},\mathbf{Y}\right)=d\left(\mathbf{Y},\mathbf{X}\right)$
(symmetry),
\item $d\left(\mathbf{X},\mathbf{Y}\right)\leq d\left(\mathbf{X},\mathbf{Z}\right)+d\left(\mathbf{Z},\mathbf{Y}\right)$
(triangle inequality).
\end{itemize}
A q-metric is a non-negative function $d\left(\cdot,\cdot\right):\Upsilon\times\Upsilon\rightarrow\left[0,\infty\right)$
that meets the identity and triangle inequality properties, but it
does not need to meet the symmetry property \cite{Wilson31,Schroeder06,Bento19}.

The triangle inequality is a key property to define distances between
points in a space, and also to measure the error of estimators. For
example, let us illustrate this with the following example, adapted
from \cite[Sec. 6.2.1]{Mahler_book14}. Let $\mathbf{X}$ be the ground
truth and let $\mathbf{Y}_{1}$ and $\mathbf{Y}_{2}$ be two estimates.
If estimate $\mathbf{Y}_{1}$ is close to $\mathbf{X}$ and estimate
$\mathbf{Y}_{2}$ is close to $\mathbf{Y}_{1}$, that is, both $d\left(\mathbf{X},\mathbf{Y}_{1}\right)$
and $d\left(\mathbf{Y}_{1},\mathbf{Y}_{2}\right)$ are small, then
$\mathbf{Y}_{2}$ should be close to $\mathbf{X}$ implying that $d\left(\mathbf{X},\mathbf{Y}_{2}\right)$
is small as well. This property is ensured for metrics and q-metrics
by the triangle inequality.

In the rest of this paper, the ground truth set of objects and the
ground truth set of trajectories are denoted by $\mathbf{x}$ and
$\mathbf{X}$. The corresponding estimates provided by a certain algorithm
are denoted by $\mathbf{y}$ and $\mathbf{Y}.$ 

\subsection{GOSPA metric\label{subsec:GOSPA-metric}}

This subsection reviews the GOSPA metric (for parameter $\alpha=2$,
as defined in \cite{Rahmathullah17}), as it is the point of reference
for the q-metrics. The ground truth sets of objects and its estimate
are written as $\ensuremath{\mathbf{x}=\left\{ x_{1},...,x_{n_{\mathbf{x}}}\right\} }$
and $\ensuremath{\mathbf{y}=\left\{ y_{1},...,y_{n_{\mathbf{y}}}\right\} }$.
The GOSPA metric looks for an optimal assignment set between the objects
in $\mathbf{x}$ and the elements in $\mathbf{y}$, with the assignment
set indicating what objects are matched, and which ones are left without
assignment. An assignment set $\theta$ meets $\theta\subseteq\left\{ 1,..,n_{\mathbf{x}}\right\} \times\left\{ 1,..,n_{\mathbf{y}}\right\} $
and, $\left(i,j\right),\left(i,j'\right)\in\theta$ implies $j=j'$
and, $\left(i,j\right),\left(i',j\right)\in\theta$ implies $i=i'$.
The set of all possible assignment sets is $\Gamma_{\mathbf{x},\mathbf{y}}$,
such that $\theta\in\Gamma_{\mathbf{x},\mathbf{y}}$. 
\begin{defn}
\label{def:GOSPA_alpha2}\textit{For a base metric $d_{b}\left(\cdot,\cdot\right)$
on the single-object space $\mathbb{X}$, a maximum localisation cost
parameter $c>0$, and a real parameter $p$ with $1\leq p<\infty$,
the GOSPA metric between sets $\mathbf{x}$ and $\mathbf{y}$ of objects
is} \cite[Prop. 1]{Rahmathullah17}
\begin{align}
 & d_{p}^{\left(c\right)}\left(\mathbf{x},\mathbf{y}\right)\nonumber \\
 & =\min_{\theta\in\Gamma_{\mathbf{x},\mathbf{y}}}\left(\sum_{\left(i,j\right)\in\theta}d_{b}\left(x_{i},y_{j}\right)^{p}+\frac{c^{p}}{2}\left(\left|\mathbf{x}\right|+\left|\mathbf{y}\right|-2\left|\theta\right|\right)\right)^{1/p}.\label{eq:GOSPA_alpha2}
\end{align}
\end{defn}
The first term in (\ref{eq:GOSPA_alpha2}) is the sum of the localisation
costs (to the $p$-th power) for pairs of assigned objects, whose
indices are $\left(i,j\right)\in\theta$. These are the objects in
$\mathbf{x}$ that have been properly detected and, therefore, have
an associated object in the estimated set $\mathbf{y}$. The terms
$\frac{c^{p}}{2}\left(\left|\mathbf{x}\right|-\left|\theta\right|\right)$
and $\frac{c^{p}}{2}\left(\left|\mathbf{y}\right|-\left|\theta\right|\right)$
are the penalties (to the $p$-th power) for the number of missed
and false objects. Each missed and each false object contributes with
a cost $\frac{c^{p}}{2}$ to the overall cost (before taking the $p$-th
root).

\section{GOSPA q-metric\label{sec:GOSPA-quasi-metric}}

This section introduces the GOSPA q-metric in Section \ref{subsec:GOSPA_q_metric_definition}.
Then, we present representative examples to illustrate how the q-metric
works in Section \ref{subsec:Examples_GOSPA_q}. The choice of parameters
is explained in Section \ref{subsec:Choice-of-parameters}. Three
properties of the GOSPA q-metric are provided in Section \ref{subsec:Properties_GOSPA_q}.

\subsection{Definition\label{subsec:GOSPA_q_metric_definition}}

The design principles (DPs) of the GOSPA q-metric are:
\begin{itemize}
\item DP1: The q-metric returns a higher value if the estimated set of targets
includes more false objects, misses more real objects or has higher
estimation errors for properly detected objects. 
\item DP2: Clear interpretability by the decomposition of the q-metric into
costs penalising the number of false objects, the number of missed
objects, and the estimation errors for properly detected objects.
\item DP3: The relative weighting of missed and false object costs can be
adapted.
\end{itemize}
DP1 and DP2 are also the DPs of the GOSPA metric. Then, the definition
of the GOSPA q-metric is the following. 
\begin{defn}
\textit{For a base q-metric $d_{b}\left(\cdot,\cdot\right)$ in the
single-object space }$\mathbb{X}$, a \textit{maximum localisation
cost parameter $c>0$, real parameter $p$ with $1\leq p<\infty$,
and q-metric parameter }$\rho\in\left(0,1\right)$,\textit{ the GOSPA
q-metric $d_{p}^{\left(c,\rho\right)}\left(\cdot,\cdot\right)$ between
two sets $\mathbf{x}$ and $\mathbf{y}$ of objects is
\begin{align}
d_{p}^{\left(c,\rho\right)}\left(\mathbf{x},\mathbf{y}\right) & =\min_{\theta\in\Gamma_{\mathbf{x},\mathbf{y}}}\left(\sum_{\left(i,j\right)\in\theta}d_{b}^{p}\left(x_{i},y_{j}\right)+\rho c^{p}\left(\left|\mathbf{y}\right|-\left|\theta\right|\right)\right.\nonumber \\
 & \quad\left.+\left(1-\rho\right)c^{p}\left(\left|\mathbf{x}\right|-\left|\theta\right|\right)\vphantom{\sum_{\left(i,j\right)\in\theta}}\right)^{1/p}.\label{eq:GOSPA_quasimetric}
\end{align}
}
\end{defn}
The differences between the GOSPA q-metric (\ref{eq:GOSPA_quasimetric})
and the GOSPA metric in (\ref{eq:GOSPA_alpha2}) are that $d_{b}\left(\cdot,\cdot\right)$
is allowed to be a q-metric, the penalty (to the $p$-th power) for
a false object is $c_{f}^{p}=\rho c^{p}$ and the penalty (to the
$p$-th power) for a missed object is $c_{m}^{p}=\left(1-\rho\right)c^{p}$,
where $\rho\in\left(0,1\right)$ is the fraction of the maximum localisation
error $c^{p}$ that represents a false object cost (to the $p$-th
power). For $\rho=1/2$ and $d_{b}\left(\cdot,\cdot\right)$ being
a metric, the GOSPA q-metric becomes the GOSPA metric. 

An alternative parameterisation of the GOSPA q-metric is to define
it in terms of $c_{f}$ and $c_{m}$, instead of $c$ and $\rho$.
The conversion between both parameterisations is straightforward.
We use the parameterisation in terms of $c$ and $\rho$ as parameter
$c$ is the maximum localisation cost that enables a pair of objects
can be assigned to each other, which is an intuitive concept. Parameter
$c$ is also used in the GOSPA metric and is used in the triangle
inequality proof.

It is direct to check the non-negativity and identity properties of
(\ref{eq:GOSPA_quasimetric}). The triangle inequality is proved in
Appendix \ref{sec:AppendixB}. The GOSPA q-metric can be computed
in polynomial time since it can be written as a 2-D assignment problem
\cite{Crouse16}.

\subsection{Examples\label{subsec:Examples_GOSPA_q}}

\begin{figure}
\begin{centering}
\includegraphics[scale=0.8]{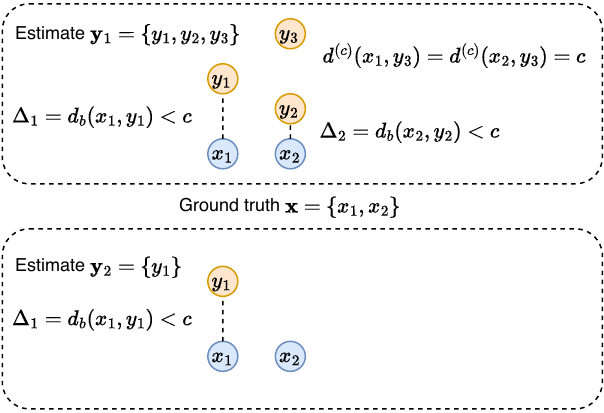}
\par\end{centering}
\caption{\label{fig:GOSPA_q_Example}Two estimated sets of objects $\mathbf{y}_{1}$
and $\mathbf{y}_{2}$ of the ground truth $\mathbf{x}.$ The dashed
lines represent assignments between the elements of the ground truth
and the estimate. Estimate $\mathbf{y}_{1}$ has two properly detected
objects with localisation errors $\Delta_{1}$ and $\Delta_{2}$ and
a false object. Estimate $\mathbf{y}_{2}$ has one properly detected
object with localisation error $\Delta_{1}$ plus a missed object.}
\end{figure}

We demonstrate the operation of the GOSPA q-metric using the examples
presented in Figure \ref{fig:GOSPA_q_Example}. For clarity, we focus
on the case where $p=1$. The GOSPA q-metric, OSPA, UOSPA, COLA and
HOTA values for the estimates $\mathbf{y}_{1}$ and $\mathbf{y}_{2}$
are given in Table \ref{tab:Distances_targets} (``Close'' columns).
For the GOSPA metric ($\rho=1/2)$, $\mathbf{y}_{2}$ is more accurate
than $\mathbf{y}_{1}$. This is because both estimates detect an object
with the same localisation error, and have either a missed or a false
object, but $\mathbf{y}_{1}$ has an additional localisation error.
For UOSPA/COLA, $\mathbf{y}_{2}$ is also more accurate than $\mathbf{y}_{1}$,
while for OSPA the ranking depends on the localisation error $\Delta_{2}$.
On the other hand, HOTA indicates that $\mathbf{y}_{1}$ is more accurate
than $\mathbf{y}_{2}$. For the GOSPA q-metric, the estimate $\mathbf{y}_{2}$
is more accurate than $\mathbf{y}_{1}$ if 
\begin{align}
\rho & >\frac{1}{2}-\frac{\Delta_{2}}{2c}.
\end{align}
That is, the parameter $\rho$, which is proportional to the false
object cost, must be sufficiently high such that $\mathbf{y}_{2}$
is considered better than $\mathbf{y}_{1}$. This is reasonable as
$\mathbf{y}_{2}$ misses an object while $\mathbf{y}_{1}$ has a false
object and one more estimate associated to a true object, so a high
enough $\rho$ makes $\mathbf{y}_{2}$ a better estimate. On the contrary,
for sufficiently small $\rho$, the GOSPA q-metric indicates that
$\mathbf{y}_{1}$ is more accurate than $\mathbf{y}_{2}$, since the
false object is penalised less. 

\begin{table}

\caption{\label{tab:Distances_targets}Distances/scores for the example in
Figure \ref{fig:GOSPA_q_Example} when some estimates are close or
all of them are far away.}

\begin{centering}
\begin{tabular}{c|cc|cc}
\hline 
 &
\multicolumn{2}{c|}{Close} &
\multicolumn{2}{c}{Far away}\tabularnewline
 &
$\mathbf{y}_{1}$ &
$\mathbf{y}_{2}$ &
$\mathbf{y}_{1}$ &
$\mathbf{y}_{2}$\tabularnewline
\hline 
GOSPA &
$\Delta_{1}+\Delta_{2}+\rho c$ &
$\Delta_{1}+\left(1-\rho\right)c$ &
$\left(2+\rho\right)c$ &
$\left(2-\rho\right)c$\tabularnewline
OSPA &
$\frac{\Delta_{1}+\Delta_{2}+c}{3}$ &
$\frac{\Delta_{1}+c}{2}$ &
$c$ &
$c$\tabularnewline
UOSPA &
$\Delta_{1}+\Delta_{2}+c$ &
$\Delta_{1}+c$ &
$3c$ &
$2c$\tabularnewline
COLA &
$\frac{\Delta_{1}+\Delta_{2}+c}{c}$ &
$\frac{\Delta_{1}+c}{c}$ &
$3$ &
$2$\tabularnewline
\hline 
HOTA($\uparrow$) &
2/3 &
1/2 &
0 &
0\tabularnewline
\hline 
\end{tabular}
\par\end{centering}
\end{table}

Let us now consider the case where $\Delta_{1}>c$ and $\Delta_{2}>c$.
That is, all estimates are far away from the real object locations.
In this case, the optimal assignments are not the ones shown in Figure
\ref{fig:GOSPA_q_Example} as it is best to leave all objects unassigned.
This implies that $\mathbf{y}_{1}$ misses two objects and declares
three false objects. On the other hand, $\mathbf{y}_{2}$ misses two
objects and contains one false object. The resulting errors are shown
in Table \ref{tab:Distances_targets} (``Far away'' columns). HOTA
and OSPA indicates that both estimates are equally good, which implies
that they do not meet DP1. GOSPA q-metric for all values of $\rho\in\left(0,1\right)$,
UOSPA and COLA indicate that $\mathbf{y}_{2}$ is more accurate than
$\mathbf{y}_{1}$, meeting DP1. An example where OSPA-UOSPA-COLA do
not meet DP1 is given in \cite[Example 2]{Rahmathullah17}.

\subsection{Choice of parameters\label{subsec:Choice-of-parameters}}

The parameters $c$ and $p$ are chosen as in the GOSPA metric. That
is, parameter $c$ is the maximum localisation error and is chosen
depending on the application. One true object state $x$ and an estimate
$y$ can only be assigned to each other if their distance according
to the base q-metric is lower than $c$. Parameter $p$ can be chosen
to adapt how outliers are penalised, with a higher value of $p$ penalising
outliers more \cite{Rahmathullah17,Schuhmacher08}. 

In most applications, the base q-metric $d_{b}\left(\cdot,\cdot\right)$
will typically be chosen to be a metric. For example, for point objects
(which do not have an extent), one can use the Euclidean distance.
 For extended objects modelled by ellipses, one can use the Gaussian
Wasserstein distance \cite{Yang16}. Parameter $\rho$ can be chosen
taking into account how much a false object should be penalised compared
to a missed object in a given application. This leads to the following
lemma.
\begin{lem}
\label{lem:Parameter_rho}If the cost of a false object (to the $p$-th
power), is $\nu\in\left(0,\infty\right)$ times the cost (to the $p$-th
power) of a missed object ($c_{f}^{p}=\nu c_{m}^{p}$), the GOSPA
q-metric parameter $\rho$ is
\begin{align}
\rho & =\frac{\nu}{\nu+1}.\label{eq:rho_choice}
\end{align}
\end{lem}
The proof of this lemma follows from the definitions of $c_{f}^{p}$
and $c_{m}^{p}$ in the paragraph after (\ref{eq:GOSPA_quasimetric}).

\subsection{Properties\label{subsec:Properties_GOSPA_q}}

This section provides several properties of the GOSPA q-metric. It
is direct to prove the following lemma.
\begin{lem}
\label{lem:GOSPA_q_symmetry}If $d_{b}\left(\cdot,\cdot\right)$ is
a metric, the GOSPA q-metric meets
\begin{align}
d_{p}^{\left(c,\rho\right)}\left(\mathbf{x},\mathbf{y}\right) & =d_{p}^{\left(c,1-\rho\right)}\left(\mathbf{y},\mathbf{x}\right).
\end{align}
\end{lem}
In addition, we prove in Appendix \ref{subsec:AppendixB_2} the following
result.
\begin{lem}
\label{lem:Optimal_assignment_GOSPA_q}The optimal assignment set
(or the optimal permutation) of the GOSPA q-metric does not depend
on $\rho$. 
\end{lem}
This result implies that for different values of $\rho$, the localisation
costs remain unchanged, but we have different missed and false object
costs, and a different overall metric value. In addition, Lemma \ref{lem:Optimal_assignment_GOSPA_q}
implies that, if $d_{b}\left(\cdot,\cdot\right)$ is a metric, the
optimal assignment set is the same as in the GOSPA metric. 

The symmetrisation property of the GOSPA q-metric is stated in the
following lemma (proved in Appendix \ref{subsec:Appendix_B_3}).
\begin{lem}
\label{lem:Symmetrisation_GOSPA}If $d_{b}\left(\cdot,\cdot\right)$
is a metric, the GOSPA metric $d_{p}^{\left(c,1/2\right)}\left(\mathbf{x},\mathbf{y}\right)$
is recovered with the following symmetrisation of the GOSPA q-metric
\begin{align}
d_{p}^{\left(c,1/2\right)}\left(\mathbf{x},\mathbf{y}\right) & =\left[\frac{1}{2}\left(d_{p}^{\left(c,\rho\right)}\left(\mathbf{x},\mathbf{y}\right)^{p}+d_{p}^{\left(c,\rho\right)}\left(\mathbf{y},\mathbf{x}\right)^{p}\right)\right]^{1/p}.\label{eq:symmetrisation_GOSPAq}
\end{align}
\end{lem}

\section{T-GOSPA q-metrics\label{sec:Trajectory-GOSPA-quasi-metric}}

This section presents the T-GOSPA q-metrics for sets of trajectories.
We first cover the GOSPA q-metric for sets of objects with at most
one element (Section \ref{subsec:Preliminaries}). We then present
the T-GOSPA q-metric for sets of trajectories in terms of multi-dimensional
assignments across time (Section \ref{subsec:Multi-dimensional-assignment-T-GOSPA}).
Then, we introduce the relaxation of the multi-dimensional assignment
T-GOSPA q-metric via linear programming (Section \ref{subsec:Linear-programming-T-GOSPA}).
Examples illustrating the T-GOSPA q-metric results are presented in
Section \ref{subsec:Examples_TGOSPA}. Finally, three properties of
the T-GOSPA q-metric are presented (Section \ref{subsec:Properties_TGOSPA_qmetric}).

\subsection{Preliminaries\label{subsec:Preliminaries}}

A building block of the T-GOSPA q-metric is the GOSPA q-metric for
sets of objects with at most one element. Given two sets of objects
$\mathbf{x}$ and $\mathbf{y}$ such that $|\mathbf{x}|\leq1$ and
$|\mathbf{y}|\leq1$, the GOSPA q-metric (\ref{eq:GOSPA_quasimetric})
becomes
\begin{align}
d_{p}^{\left(c,\rho\right)}\left(\mathbf{x},\mathbf{y}\right) & \triangleq\begin{cases}
\min\left(c,d_{b}\left(x,y\right)\right) & \mathbf{x}=\left\{ x\right\} ,\mathbf{y}=\left\{ y\right\} \\
\rho^{1/p}c & \mathbf{x}=\emptyset,\mathbf{y}=\left\{ y\right\} \\
\left(1-\rho\right)^{1/p}c & \mathbf{x}=\{x\},\mathbf{y}=\emptyset\\
0 & \mathbf{x}=\mathbf{y}=\emptyset.
\end{cases}\label{eq:baseQMetric}
\end{align}

\subsection{Multi-dimensional assignment T-GOSPA q-metric\label{subsec:Multi-dimensional-assignment-T-GOSPA}}

This section presents the T-GOSPA q-metric as the solution to a multi-dimensional
assignment problem. The DPs of the T-GOSPA q-metric are DP3 and 
\begin{itemize}
\item DP4: The q-metric gives a higher value when the estimated set of trajectories
includes more false objects, misses more real objects, has higher
estimation errors for properly detected objects or has higher number
of track switches. 
\item DP5: Clear interpretability by the decomposition of the q-metric into
costs penalising the number of false objects, the number of missed
objects, the number of track switches, and the estimation errors for
properly detected objects.
\end{itemize}
DP4 and DP5 are the same DPs as of the T-GOSPA metric. The ground
truth sets of trajectories and its estimate are written as $\ensuremath{\mathbf{X}=\left\{ X_{1},...,X_{n_{\mathbf{X}}}\right\} }$
and $\ensuremath{\mathbf{Y}=\left\{ Y_{1},...,Y_{n_{\mathbf{Y}}}\right\} }$.
For the trajectories $X_{i}$ and $Y_{j}$, the corresponding set
of objects at time step $k$ are represented by $\mathbf{x}_{i}^{k}$
and $\mathbf{y}_{j}^{k}$. It is met that $\left|\mathbf{x}_{i}^{k}\right|\leq1$
and $\left|\mathbf{y}_{j}^{k}\right|\leq1$ as these sets are either
empty, if the corresponding trajectory does not have an object state
at time step $k$, or have a single element. The number of objects
in $\mathbf{X}$ that are present at time step $k$ is denoted $n_{\mathbf{X}}^{k}=\sum_{i=1}^{n_{\mathbf{X}}}\left|\mathbf{x}_{i}^{k}\right|$.
A similar notation, $n_{\mathbf{Y}}^{k}$, is used for $\mathbf{Y}$. 

In the trajectory q-metrics, we make assignments at each time step
between the trajectories in $\mathbf{X}$ and those in $\mathbf{Y}$.
That is, at each time step, each set $\mathbf{x}_{i}^{k}$ is associated
to a set $\mathbf{y}_{j}^{k}$ or is left without an assignment. 

To represent these assignments, we can use assignment vectors as follows.
An assignment vector at time step $k$ is written as $[\pi_{1}^{k},...,\pi_{n_{\mathbf{X}}}^{k}]$,
with $n_{\mathbf{X}}$ being its length. If its $i$-th entry $\pi_{i}^{k}=j$,
with $j\in\{1,\ldots,n_{\mathbf{Y}}\}$, it means that $\mathbf{x}_{i}^{k}$
is assigned to $\mathbf{y}_{j}^{k}$. As there can be at maximum one
$\mathbf{x}_{i}^{k}$ assigned to $\mathbf{y}_{j}^{k}$, and the other
way round, the assignment vector has the constraint that $\pi_{i}^{k}=\pi_{i^{\prime}}^{k}=j>0$
implies $i=i^{\prime}$. A value $\pi_{i}^{k}=0$ implies that $\mathbf{x}_{i}^{k}$
is unassigned. The set of all these possible assignment vectors is
then written as $\ensuremath{\Pi_{\mathbf{X},\mathbf{Y}}}.$

\begin{defn}
\label{def:MD-trajectory-quasimetric} \textit{Given a base q-metric
$d_{b}\left(\cdot,\cdot\right)$ in the single-object space $\mathbb{X}$,
a maximum localisation cost $c>0$, a real parameter $p$ with $1\leq p<\infty$,
a track switching penalty $\gamma>0$, and q-metric parameter} $\rho\in\left(0,1\right)$\textit{,
the T-GOSPA q-metric $d_{p}^{\left(c,\rho,\gamma\right)}\left(\mathbf{X},\mathbf{Y}\right)$
between two sets $\mathbf{X}$ and $\mathbf{Y}$ of trajectories is}
\begin{align}
d_{p}^{\left(c,\rho,\gamma\right)}\left(\mathbf{X},\mathbf{Y}\right) & =\min_{\substack{\pi^{k}\in\Pi_{\mathbf{X},\mathbf{Y}}\\
k=1,\ldots,T
}
}\left(\sum_{k=1}^{T}d_{\mathbf{X},\mathbf{Y}}^{k}\left(\mathbf{X},\mathbf{Y},\pi^{k}\right)^{p}\right.\nonumber \\
 & \:\left.+\sum_{k=1}^{T-1}s_{\mathbf{X},\mathbf{Y}}\left(\pi^{k},\pi^{k+1}\right)^{p}\right)^{1/p}\label{eq:multi-dimensional-quasimetric}
\end{align}
\textit{where}
\begin{align}
d_{\mathbf{X},\mathbf{Y}}^{k}\left(\mathbf{X},\mathbf{Y},\pi^{k}\right)^{p} & =\sum_{\left(i,j\right)\in\theta^{k}\left(\pi^{k}\right)}d_{p}^{\left(c,\rho\right)}\left(\mathbf{x}_{i}^{k},\mathbf{y}_{j}^{k}\right)^{p}\nonumber \\
 & +\rho c^{p}\left(n_{\mathbf{Y}}^{k}-\left|\theta^{k}\left(\pi^{k}\right)\right|\right)\nonumber \\
 & +\left(1-\rho\right)c^{p}\left(n_{\mathbf{X}}^{k}-\left|\theta^{k}\left(\pi^{k}\right)\right|\right)\label{eq:GOSPA_no_minimisation}
\end{align}
\textit{includes the costs (to the $p$-th power) for properly detected
objects (first line), false objects (second line) and missed objects
(third line) at time step $k$, $d_{p}^{\left(c,\rho\right)}\left(\cdot,\cdot\right)$
is the GOSPA q-metric in (\ref{eq:baseQMetric}),}
\begin{align}
\theta^{k}\left(\pi^{k}\right) & =\left\{ (i,\pi_{i}^{k}):i\in\{1,\ldots,n_{\mathbf{X}}\},\vphantom{d_{p}^{\left(c,\rho\right)}\left(\mathbf{x}_{i}^{k},\mathbf{y}_{\pi_{i}^{k}}^{k}\right)}\right.\nonumber \\
 & \,\left.\left|\mathbf{x}_{i}^{k}\right|=\left|\mathbf{y}_{\pi_{i}^{k}}^{k}\right|=1,\,d_{p}^{\left(c,\rho\right)}\left(\mathbf{x}_{i}^{k},\mathbf{y}_{\pi_{i}^{k}}^{k}\right)<c\right\} \label{eq:target_level_assignment}
\end{align}
\textit{and the track switch penalty (to the $p$-th power) between
time step $k$ and $k+1$ is}
\begin{equation}
s_{\mathbf{X},\mathbf{Y}}(\pi^{k},\pi^{k+1})^{p}=\gamma^{p}\sum_{i=1}^{n_{\mathbf{X}}}s\big(\pi_{i}^{k},\pi_{i}^{k+1}\big)\label{eq:switching_cost_MD}
\end{equation}
\begin{align*}
s\big(\pi_{i}^{k},\pi_{i}^{k+1}\big) & =\begin{cases}
0 & \pi_{i}^{k}=\pi_{i}^{k+1}\\
1 & \pi_{i}^{k}\neq\pi_{i}^{k+1},\pi_{i}^{k}\neq0,\pi_{i}^{k+1}\neq0\\
\frac{1}{2} & \text{otherwise. \quad\quad\quad\quad\quad\quad\quad\quad}
\end{cases}
\end{align*}
\end{defn}
The proof that $d_{p}^{\left(c,\rho,\gamma\right)}\left(\cdot,\cdot\right)$
in Definition \ref{def:MD-trajectory-quasimetric} is a q-metric,
meeting the properties indicated in Section \ref{subsec:Metric-and-q-metrics},
 will be included as part of the proof for its linear programming
version in the next subsection. The T-GOSPA q-metric above (and also
the linear programming relaxation version in the next subsection)
are simply using the GOSPA q-metric (\ref{eq:baseQMetric}) inside
the original T-GOSPA metric in \cite{Angel20_d}. The T-GOSPA metric
is obtained by setting $\rho=1/2$ provided that $d_{b}\left(\cdot,\cdot\right)$
is a metric. Apart from the costs for false, missed and localization
errors (already present in the GOSPA q-metric), the T-GOSPA q-metric
also introduces a cost for track switches that remains symmetric in
the q-metric, to enable the linear programming relaxation. Equation
(\ref{eq:GOSPA_no_minimisation}) corresponds to the GOSPA q-metric,
but with the object-level assignment set $\theta^{k}\left(\pi^{k}\right)$
being determined by the trajectory-level association $\pi^{k}$. For
two objects to be assigned in $\theta^{k}\left(\pi^{k}\right)$, their
base q-metric must be smaller than $c$, as indicated by (\ref{eq:target_level_assignment}).
For one time step $T=1$, the T-GOSPA q-metric becomes the GOSPA q-metric.
Parameter $\rho$ in the T-GOSPA q-metric can be chosen as in the
GOSPA q-metric, see Section \ref{subsec:Choice-of-parameters}. The
rest of the parameters can be chosen as in the T-GOSPA metric \cite{Angel20_d}. 

\subsection{Linear programming T-GOSPA q-metric\label{subsec:Linear-programming-T-GOSPA}}

To define the linear programming T-GOSPA q-metric, we first need to
provide some additional notation. The transpose of a matrix $A$ is
written as $A^{\dagger}$. The assignments of the T-GOSPA q-metric
can be written using binary matrices \cite{Angel20_d}. A matrix $W^{k}$
of size $(n_{\mathbf{X}}+1)\times(n_{\mathbf{Y}}+1)$ that represents
assignments between $\mathbf{X}$ and $\mathbf{Y}$ meets the properties:

\begin{align}
W^{k}(i,j) & \in\{0,1\},\forall\ i,j\label{eq:binary_constraint4_rearranged}\\
\sum_{i=1}^{n_{\mathbf{X}}+1}W^{k}(i,j) & =1,\ j=1,\ldots,n_{\mathbf{Y}}\label{eq:binary_constraint1}\\
\sum_{j=1}^{n_{\mathbf{Y}}+1}W^{k}(i,j) & =1,\ i=1,\ldots,n_{\mathbf{X}}\label{eq:binary_constraint2}\\
W^{k}(n_{\mathbf{X}}+1,n_{\mathbf{Y}}+1) & =0,\label{eq:binary_constraint3}
\end{align}
where $W^{k}(i,j)$ is the $(i,j)$ element of matrix $W^{k}$. A
value $W^{k}(i,j)=1$ means that $\mathbf{x}_{i}^{k}$ is assigned
to $\mathbf{y}_{j}^{k}$ and a value $W^{k}(i,j)=0$ means that $\mathbf{x}_{i}^{k}$
is not assigned to $\mathbf{y}_{j}^{k}$. In addition, if $\mathbf{x}_{i}^{k}$
is left without assignment, $W^{k}(i,n_{\mathbf{Y}}+1)=1$. Similarly,
if $\mathbf{y}_{j}^{k}$ is left without assignment, $W^{k}(n_{\mathbf{X}}+1,j)=1$.
The set of binary matrices that meet (\ref{eq:binary_constraint4_rearranged})-(\ref{eq:binary_constraint3})
is $\mathcal{W}_{\mathbf{X},\mathbf{Y}}$. 

The linear programming T-GOSPA q-metric is based on changing the binary
constraint (\ref{eq:binary_constraint4_rearranged}) by a relaxed
version
\begin{align}
W^{k}(i,j)\geq0, & \quad\forall i,j.\label{eq:LP_constraint}
\end{align}
The set of matrices that meet (\ref{eq:binary_constraint1})-(\ref{eq:binary_constraint3})
and (\ref{eq:LP_constraint}) is denoted as $\mathcal{\overline{W}}_{\mathbf{X},\mathbf{Y}}$. 
\begin{prop}
\label{prop:LP_trajectory-quasimetric}Given a base q-metric $d_{b}\left(\cdot,\cdot\right)$
in the single-object space $\mathbb{X}$, a maximum localisation cost
$c>0$, a real parameter $p$ with $1\leq p<\infty$, a track switching
penalty $\gamma>0$, and q-metric parameter $\rho\in\left(0,1\right)$,
the LP relaxation $\overline{d}_{p}^{\left(c,\rho,\gamma\right)}\left(\mathbf{X},\mathbf{Y}\right)$
of $d_{p}^{\left(c,\rho,\gamma\right)}\left(\mathbf{X},\mathbf{Y}\right)$
is a q-metric with expression
\begin{align}
\overline{d}_{p}^{\left(c,\rho,\gamma\right)}\left(\mathbf{X},\mathbf{Y}\right) & =\min_{\substack{W^{k}\in\mathcal{\overline{W}}_{\mathbf{X},\mathbf{Y}}\\
k=1,\ldots,T
}
}\Bigg(\sum_{k=1}^{T}\mathrm{tr}\big[\big(D_{\mathbf{X},\mathbf{Y}}^{k}\big)^{\dagger}W^{k}\big]\nonumber \\
 & +\frac{\gamma^{p}}{2}\sum_{k=1}^{T-1}\sum_{i=1}^{n_{\mathbf{X}}}\sum_{j=1}^{n_{\mathbf{Y}}}|W^{k}(i,j)-W^{k+1}(i,j)|\Bigg)^{\frac{1}{p}},\label{eq:LP_quasimetric}
\end{align}
where $D_{\mathbf{X},\mathbf{Y}}^{k}$ is a matrix of size $(n_{\mathbf{X}}+1)\times(n_{\mathbf{Y}}+1)$
whose $(i,j)$ element is
\begin{align}
D_{\mathbf{X},\mathbf{Y}}^{k}(i,j) & =d_{p}^{\left(c,\rho\right)}\left(\mathbf{x}_{i}^{k},\mathbf{y}_{j}^{k}\right)^{p}\label{eq:D_i_j}
\end{align}
where we recall that $d_{p}^{\left(c,\rho\right)}\left(\cdot,\cdot\right)$
is the GOSPA q-metric (\ref{eq:baseQMetric}), and $\mathbf{x}_{n_{\mathbf{X}}+1}^{k}=\emptyset$
and $\mathbf{y}_{n_{\mathbf{Y}}+1}^{k}=\emptyset$. 
\end{prop}
The identity property of the q-metric is direct and the triangle
inequality is proved in Appendix \ref{sec:AppendixD}. It should be
noted that if instead of optimising over $\mathcal{\overline{W}}_{\mathbf{X},\mathbf{Y}}$
in (\ref{eq:LP_quasimetric}), we optimise over $\mathcal{W}_{\mathbf{X},\mathbf{Y}}$
(the non-relaxed assignment matrices), (\ref{eq:LP_quasimetric})
becomes (\ref{eq:multi-dimensional-quasimetric}). In addition, as
in \cite{Angel21_f}, it is possible to add time weights in both (\ref{eq:multi-dimensional-quasimetric})
and (\ref{eq:LP_quasimetric}) and the q-metric properties are preserved.
The T-GOSPA q-metric (LP implementation) can be computed in polynomial
time since it is an LP \cite{Khachiyan80}. Clustering can also be
used to speed up computation \cite[Sec. IV.D]{Angel20_d}. It is also
possible to obtain approximate, fast computations following the approaches
in \cite{Wernholm25,Warnsater25}.

As the T-GOSPA metric, the T-GOSPA q-metric can be decomposed into
different costs associated to localisation errors, missed objects,
false objects and track switches \cite[Sec. IV.C]{Angel20_d}. This
decomposition provides a clear interpretability of the obtained results.
It is also direct to apply the q-metric with trajectories with gaps
\cite{Angel20_d}.

\subsection{Examples\label{subsec:Examples_TGOSPA}}

We illustrate how the T-GOSPA q-metric works in two examples.

\subsubsection{Example 1}

\begin{figure}
\begin{centering}
\includegraphics[scale=0.8]{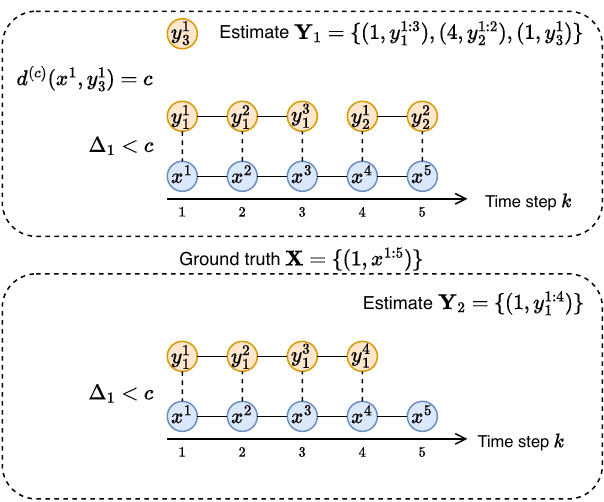}
\par\end{centering}
\caption{\label{fig:TGOSPA_q_Example}Two estimated sets of trajectories $\mathbf{Y}_{1}$
and $\mathbf{Y}_{2}$ of the ground truth $\mathbf{X}.$ The dashed
lines represent assignments between the elements of the ground truth
and the estimate. Estimate $\mathbf{Y}_{1}$ has five correct detections
with localisation error $\Delta_{1}$, a false object and a track
switch. Estimate $\mathbf{Y}_{2}$ has four properly detected objects
with localisation error $\Delta_{1}$ and a missed object.}
\end{figure}

We illustrate how the T-GOSPA q-metric works under DP3-DP5 for the
example in Figure \ref{fig:TGOSPA_q_Example}. As before, we consider
the case $p=1$. For sufficiently small $\gamma$, estimate $\mathbf{Y}_{1}$
has five properly detected objects with localisation error $\Delta_{1}$,
a false object and a track switch, resulting in the value
\begin{align}
d_{1}^{\left(c,\rho\right)}\left(\mathbf{X},\mathbf{Y}_{1}\right) & =5\Delta_{1}+\rho c+\gamma.\label{eq:d_X_Y1_TGOSPA}
\end{align}
It should be noted that, if $\gamma$ were sufficiently high, the
optimal assignments are fixed across time, and therefore the optimal
assignment for $\mathbf{Y}_{1}$ would not be the one in Figure \ref{fig:TGOSPA_q_Example}.
In this case, it would be optimal to leave the trajectory $\left(4,y_{2}^{1:2}\right)$
unassigned with an overall q-metric value $3\Delta_{1}+\rho c+2c$.
That is, we substitute the localisation costs for two objects $\left(2\Delta_{1}\right)$
plus the track switching cost $\gamma$ in (\ref{eq:d_X_Y1_TGOSPA})
by the cost of two missed objects and two false objects ($2c$), which
would result in $\mathbf{Y}_{2}$ being always better than $\mathbf{Y}_{1}$,
regardless of $\rho$.

Therefore, for sufficiently small $\gamma$, the optimal assignment
is the one in Figure \ref{fig:TGOSPA_q_Example} if 
\begin{align}
5\Delta_{1}+\rho c+\gamma & <3\Delta_{1}+\rho c+2c
\end{align}
which implies that the track switching cost must meet
\begin{align}
\gamma & <2\left(c-\Delta_{1}\right).
\end{align}

On the other hand, estimate $\mathbf{Y}_{2}$ has four properly detected
objects with localisation error $\Delta_{1}$ and a missed object,
resulting in the value
\begin{align}
d_{1}^{\left(c,\rho\right)}\left(\mathbf{X},\mathbf{Y}_{2}\right) & =4\Delta_{1}+\left(1-\rho\right)c.
\end{align}
According to the T-GOSPA q-metric, the estimate $\mathbf{Y}_{2}$
is more accurate than $\mathbf{Y}_{1}$ if 
\begin{align}
\rho & >\frac{c-\Delta_{1}-\gamma}{2c}.
\end{align}
  The values of OSPA$^{(2)}$, UOSPA$^{(2)}$ and HOTA values are
given in Table \ref{tab:Distances_trajectories1} (``Close estimates''
column). The T-GOSPA metric ($\rho=0.5$), OSPA$^{(2)}$, UOSPA$^{(2)}$
and HOTA indicate that $\mathbf{Y}_{2}$ is more accurate than $\mathbf{Y}_{1}$.
However, setting $\rho$ small enough (indicating a low cost for false
objects), the T-GOSPA q-metric can determine that $\mathbf{Y}_{1}$
is more accurate. For instance, setting $c=1$, $\Delta_{1}=0.1$
and $\gamma=0.1$, $\mathbf{Y}_{1}$ is more accurate than $\mathbf{Y}_{2}$
if $\rho<0.4$.

Let us now analyse the case where all estimates are far away ($\Delta_{1}>c$)
in Figure \ref{tab:Distances_trajectories2}. In this case, $\mathbf{Y}_{1}$
has 5 missed objects and 6 false objects, while $\mathbf{Y}_{2}$
has 5 missed objects and 4 false objects. T-GOSPA and UOSPA$^{(2)}$
indicate that $\mathbf{Y}_{2}$ is more accurate, meeting DP4. On
the contrary, OSPA$^{(2)}$ and HOTA indicate that both estimates
are equally accurate.

\begin{table}
\caption{\label{tab:Distances_trajectories1}Distances/scores for the example
in Figure \ref{fig:TGOSPA_q_Example} when some estimates are close
and when all of them are far away.}

\centering{}%
\begin{tabular}{c|cc|cc}
\hline 
 &
\multicolumn{2}{c|}{Close } &
\multicolumn{2}{c}{Far away }\tabularnewline
 &
$\mathbf{Y}_{1}$ &
$\mathbf{Y}_{2}$ &
$\mathbf{Y}_{1}$ &
$\mathbf{Y}_{2}$\tabularnewline
\hline 
TGOSPA &
\begin{cellvarwidth}[t]
\centering

$\begin{array}{c}
5\Delta_{1}+\rho c\\
+\gamma
\end{array}$
\end{cellvarwidth} &
$\begin{array}{c}
4\Delta_{1}\\
+\left(1-\rho\right)c
\end{array}$ &
$\left(5+\rho\right)c$ &
$\left(5-\rho\right)c$\tabularnewline
OSPA$^{(2)}$ &
$\frac{\Delta_{1}+4c}{5}$ &
$\frac{4\Delta_{1}+c}{5}$ &
$c$ &
$c$\tabularnewline
UOSPA$^{(2)}$ &
$\frac{3\Delta_{1}+12c}{5}$ &
$\frac{4\Delta_{1}+c}{5}$ &
$3c$ &
$c$\tabularnewline
\hline 
HOTA($\uparrow$) &
$\sqrt{\frac{13}{30}}\approx0.66$ &
$\frac{4}{5}=0.8$ &
0 &
0\tabularnewline
\hline 
\end{tabular}
\end{table}

\subsubsection{Example 2}

\begin{figure}
\begin{centering}
\includegraphics[scale=0.8]{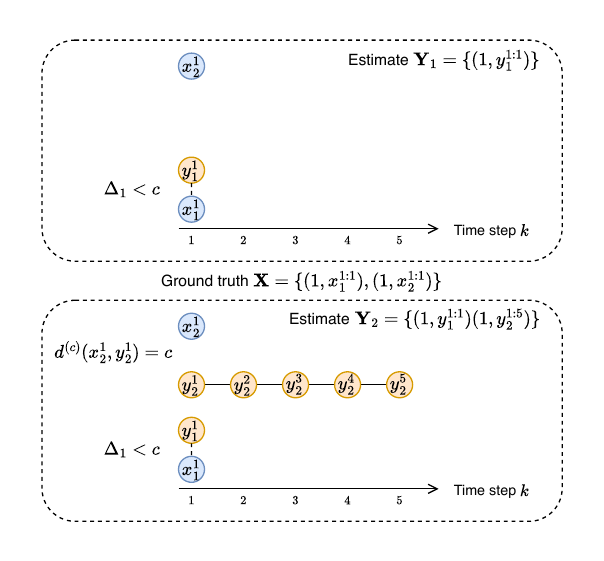}
\par\end{centering}
\caption{\label{fig:TGOSPA_q_Example2}Two estimated sets of trajectories $\mathbf{Y}_{1}$
and $\mathbf{Y}_{2}$ of the ground truth $\mathbf{X}.$ The dashed
lines represent assignments between the elements of the ground truth
and the estimate. Estimate $\mathbf{Y}_{1}$ has five one correct
detection with localisation error $\Delta_{1}$, and a missed object.
Estimate $\mathbf{Y}_{2}$ has an additional false trajectory of length
5.}
\end{figure}

We now consider the example in Figure \ref{fig:TGOSPA_q_Example2},
in which estimate $\mathbf{Y}_{2}$ has an extra false trajectory
of length 5 compared to $\mathbf{Y}_{1}$. The errors are shown in
Table \ref{tab:Distances_trajectories2}. OSPA$^{(2)}$ and UOSPA$^{(2)}$
do not penalise this extra false trajectory (which could be arbitrarily
long) and indicate that $\mathbf{Y}_{1}$ and $\mathbf{Y}_{2}$ are
equally accurate. HOTA and T-GOSPA work according to DP4. If all estimates
are far away, only T-GOSPA indicates that $\mathbf{Y}_{1}$ is more
accurate than $\mathbf{Y}_{2}$, as expected according to DP4.

\begin{table}
\caption{\label{tab:Distances_trajectories2}Distances/scores for the example
in Figure \ref{fig:TGOSPA_q_Example2} when some estimates are close
and when all of them are far away.}

\centering{}%
\begin{tabular}{>{\centering}p{1.1cm}|>{\centering}p{1.2cm}c|cc}
\hline 
 &
\multicolumn{2}{c|}{Close} &
\multicolumn{2}{c}{Far away}\tabularnewline
 &
$\mathbf{Y}_{1}$ &
$\mathbf{Y}_{2}$ &
$\mathbf{Y}_{1}$ &
$\mathbf{Y}_{2}$\tabularnewline
\hline 
TGOSPA &
$\begin{array}{c}
\Delta_{1}\\
+\left(1-\rho\right)c
\end{array}$ &
$\begin{array}{c}
\Delta_{1}\\
+\left(1+4\rho\right)c
\end{array}$ &
$\left(2-\rho\right)c$ &
$\left(2+4\rho\right)c$\tabularnewline
OSPA$^{(2)}$ &
$\frac{\Delta_{1}+c}{2}$ &
$\frac{\Delta_{1}+c}{2}$ &
$c$ &
$c$\tabularnewline
UOSPA$^{(2)}$ &
$\Delta_{1}+c$ &
$\Delta_{1}+c$ &
$2c$ &
$2c$\tabularnewline
\hline 
HOTA($\uparrow$) &
$\frac{1}{\sqrt{2}}$ &
$\frac{1}{\sqrt{7}}$ &
0 &
0\tabularnewline
\hline 
\end{tabular}
\end{table}

\subsection{Properties\label{subsec:Properties_TGOSPA_qmetric}}

This section extends the properties of the GOSPA q-metric presented
in Section \ref{subsec:Properties_GOSPA_q} to the T-GOSPA q-metric.
The following result holds directly from the definition.
\begin{lem}
\label{lem:TGOSPA_q_symmetry}If $d_{b}\left(\cdot,\cdot\right)$
is a metric, the T-GOSPA q-metric and its LP relaxation meet
\begin{align}
d_{p}^{\left(c,\rho,\gamma\right)}\left(\mathbf{X},\mathbf{Y}\right) & =d_{p}^{\left(c,1-\rho,\gamma\right)}\left(\mathbf{Y},\mathbf{X}\right)\\
\overline{d}_{p}^{\left(c,\rho,\gamma\right)}\left(\mathbf{X},\mathbf{Y}\right) & =\overline{d}_{p}^{\left(c,1-\rho,\gamma\right)}\left(\mathbf{Y},\mathbf{X}\right).
\end{align}
\end{lem}
The following lemma regarding the optimal assignment is proved in
Appendix \ref{subsec:Appendix_D1}.
\begin{lem}
\label{lem:Optimal_assignment_TGOSPA_q}The optimal sequence of matrices
$W^{k}\in\mathcal{\overline{W}}_{\mathbf{X},\mathbf{Y}}$ with $k=1,\ldots,T,$
in the LP T-GOSPA q-metric does not depend on $\rho$. Similarly,
the optimal sequence of matrices $W^{k}\in\mathcal{W}_{\mathbf{X},\mathbf{Y}}$
with $k=1,\ldots,T$ in the T-GOSPA q-metric does not depend on $\rho$. 
\end{lem}
A direct consequence of this lemma is that the optimal sequence of
assignment vectors $\pi^{k}\in\Pi_{\mathbf{X},\mathbf{Y}},\,k=1,\ldots,T$
of the T-GOSPA q-metric does not depend on $\rho$. Lemma \ref{lem:Optimal_assignment_TGOSPA_q}
implies that for different values of $\rho$, the localisation and
track switching costs remain unchanged, but there are different missed
and false object costs. In addition, this implies that, if $d_{b}\left(\cdot,\cdot\right)$
is a metric, the optimal $W^{k}$ is the same as in the T-GOSPA metric.
If we want to compute the q-metric for several values of $\rho$,
this lemma provides a computational advantage, as it suffices to solve
the optimisation problem once. 

The T-GOSPA q-metric symmetrisation property is provided in the following
lemma, proved in Appendix \ref{subsec:Appendix_D2}.
\begin{lem}
\label{lem:Symmetrisation_TGOSPA}If $d_{b}\left(\cdot,\cdot\right)$
is a metric, the T-GOSPA metrics $d_{p}^{\left(c,1/2,\gamma\right)}\left(\mathbf{X},\mathbf{Y}\right)$
and $\overline{d}_{p}^{\left(c,1/2,\gamma\right)}\left(\mathbf{X},\mathbf{Y}\right)$
are recovered with the following symmetrisation of the TGOSPA q-metrics
\begin{align}
 & d_{p}^{\left(c,1/2,\gamma\right)}\left(\mathbf{X},\mathbf{Y}\right)\nonumber \\
 & \quad=\left[\frac{1}{2}\left(d_{p}^{\left(c,\rho,\gamma\right)}\left(\mathbf{X},\mathbf{Y}\right)^{p}+d_{p}^{\left(c,\rho,\gamma\right)}\left(\mathbf{Y},\mathbf{X}\right)^{p}\right)\right]^{1/p},\label{eq:symmetrisation_TGOSPAq_1}\\
 & \overline{d}_{p}^{\left(c,1/2,\gamma\right)}\left(\mathbf{X},\mathbf{Y}\right)\nonumber \\
 & \quad=\left[\frac{1}{2}\left(\overline{d}_{p}^{\left(c,\rho,\gamma\right)}\left(\mathbf{X},\mathbf{Y}\right)^{p}+\overline{d}_{p}^{\left(c,\rho,\gamma\right)}\left(\mathbf{Y},\mathbf{X}\right)^{p}\right)\right]^{1/p}.\label{eq:symmetrisation_TGOSPAq_2}
\end{align}
\end{lem}

\section{Q-metric-based similarity score functions\label{sec:Q-metric-based-scores}}

In this section, we present the definition of q-metric-based scores
in Section \ref{subsec:Q-metric-based-score-}. Then, in Section \ref{subsec:Examples-metric-preserving-mappings},
we provide examples of metric-preserving mappings, which are required
to define the q-metric-based scores. Section \ref{subsec:Score_Application-to-GOSPA_q}
defines the q-metric-based scores based on GOSPA and T-GOSPA.

\subsection{Q-metric-based score definition\label{subsec:Q-metric-based-score-}}

We define a q-metric-based score function on a given space $\Upsilon$,
to measure the similarity between $\mathbf{X},\mathbf{Y}\in\Upsilon$,
as a function $s\left(\cdot,\cdot\right):\Upsilon\times\Upsilon\rightarrow\left[0,1\right]$
that meets the properties
\begin{itemize}
\item $s\left(\mathbf{X},\mathbf{Y}\right)=1$ if and only if $\mathbf{X}=\mathbf{Y}$,
\item $s\left(\mathbf{X},\mathbf{Y}\right)+1\geq s\left(\mathbf{X},\mathbf{Z}\right)+s\left(\mathbf{Z},\mathbf{Y}\right)$.
\end{itemize}
If, additionally, $s\left(\mathbf{X},\mathbf{Y}\right)=s\left(\mathbf{Y},\mathbf{X}\right)$,
we say that $s\left(\cdot,\cdot\right)$ is a metric-based score.
Clearly, $s\left(\cdot,\cdot\right)$ is a q-metric-based score if
and only if $1-s\left(\cdot,\cdot\right)$ is a q-metric. These properties
define score functions based on mathematically principled notions
of error (given by metrics or q-metrics). We consider two cases to
obtain q-metric-based scores, the first one is for bounded q-metrics,
and the second one is for unbounded q-metrics.

Let us consider $d\left(\cdot,\cdot\right)$ is a bounded q-metric
(for instance, the OSPA metric \cite{Schuhmacher08_b,Schuhmacher08}),
whose image supremum is $c$. Then, a q-metric-based score is simply
obtained as $1-d\left(\cdot,\cdot\right)/c$. Now, let us proceed
to build score functions for unbounded q-metrics (such as the ones
proposed in this paper, UOSPA or COLA) using metric-preserving mappings
\cite{Corazza99}. 

We first consider a function $f\left(\cdot\right):\left[0,\infty\right)\rightarrow\left[0,1\right]$
with the properties:
\begin{itemize}
\item P1: $f^{-1}\left(0\right)=\{0\}$. 
\item P2: $f\left(\cdot\right)$ is non-decreasing. 
\item P3: $f\left(\cdot\right)$ is sub-additive, meaning that \cite[Chap. VII]{Hille_book57},
for $a\geq0$ and $b\geq0$,
\begin{align}
f\left(a+b\right) & \leq f\left(a\right)+f\left(b\right).
\end{align}
\item P4: $\lim_{x\rightarrow\infty}f\left(x\right)=1$. 
\end{itemize}
P1-P3 implies that $f\left(\cdot\right)$ is a metric-preserving mapping,
and P4 is required to build a score function. 
\begin{lem}
\label{lem:Bounding_metric}Given a function $f\left(\cdot\right)$
that meets P1-P4 and a q-metric $d\left(\cdot,\cdot\right)$ on a
space $\Upsilon$, their composition , $f\left(d\left(\cdot,\cdot\right)\right)$
is a q-metric on space $\Upsilon$ taking values in $\left[0,1\right]$.
This implies that $s\left(\cdot,\cdot\right)=1-f\left(d\left(\cdot,\cdot\right)\right)$
is a q-metric-based similarity score function.
\end{lem}
For completeness, the proof of Lemma \ref{lem:Bounding_metric} is
provided in Appendix \ref{sec:Proof_metric_preserving_mapping}. It
should be noted that P1-P3 imply that $f\left(\cdot\right)$ is a
metric-preserving mapping \cite{Corazza99} and are required to keep
the q-metric properties. P4 is required such that the composed q-metric
is bounded with supremum equal to 1, and therefore $1-f\left(d\left(\cdot,\cdot\right)\right)$
is a q-metric-based score. 

\subsection{Examples of metric-preserving mappings\label{subsec:Examples-metric-preserving-mappings}}

This section provides some examples of functions that meet P1-P4 and
define q-metric-based scores for unbounded q-metrics. We include a
positive parameter $\beta>0$ to the mappings to give more flexibility
to the design of the score. The mappings are the scaled and translated
sigmoid (\ref{eq:metric_preserving1}), the hyperbolic tangent function
(\ref{eq:metric_preserving2}), the scaled arctangent function (\ref{eq:metric_preserving3})
and a fractional linear function (\ref{eq:metric_preserving4}):
\begin{align}
f_{\beta}\left(x\right) & =2\frac{1}{1+e^{-x/\beta}}-1\label{eq:metric_preserving1}\\
f_{\beta}\left(x\right) & =\tanh\left(x/\beta\right)\label{eq:metric_preserving2}\\
f_{\beta}\left(x\right) & =\frac{2}{\pi}\arctan\left(x/\beta\right)\label{eq:metric_preserving3}\\
f_{\beta}\left(x\right) & =\frac{x/\beta}{1+x/\beta}.\label{eq:metric_preserving4}
\end{align}
These functions achieve keep the metric properties and map unbounded
metric values to the interval $\left[0,1\right]$, with slightly different
ways to perform the mapping.

\subsection{Application to GOSPA and T-GOSPA q-metrics\label{subsec:Score_Application-to-GOSPA_q}}

Applying Lemma \ref{lem:Bounding_metric} and the metric-preserving
mappings (\ref{eq:metric_preserving1})-(\ref{eq:metric_preserving4})
to the GOSPA and T-GOSPA q-metrics, we obtain bounded versions of
GOSPA and T-GOSPA q-metrics. If $f_{\beta}\left(\cdot\right)$ is
an increasing function (which meets P2), then the bounded versions
also meet the design principles DP1-DP3 or DP3-DP5, with the decomposition
applying before the mapping $f_{\beta}\left(\cdot\right)$. In this
case, we can see that, while an increase of false objects is always
penalised, each additional false object is penalised less. In addition,
given sets $\mathbf{X}$, $\mathbf{Y}$ and $\mathbf{Z}$, if 
\begin{align}
d_{p}^{\left(c,\rho,\gamma\right)}\left(\mathbf{X},\mathbf{Y}\right) & <d_{p}^{\left(c,\rho,\gamma\right)}\left(\mathbf{X},\mathbf{Z}\right),
\end{align}
then, 
\begin{align}
f_{\beta}\left(d_{p}^{\left(c,\rho,\gamma\right)}\left(\mathbf{X},\mathbf{Y}\right)\right) & <f_{\beta}\left(d_{p}^{\left(c,\rho,\gamma\right)}\left(\mathbf{X},\mathbf{Z}\right)\right)\\
1-f_{\beta}\left(d_{p}^{\left(c,\rho,\gamma\right)}\left(\mathbf{X},\mathbf{Y}\right)\right) & >1-f_{\beta}\left(d_{p}^{\left(c,\rho,\gamma\right)}\left(\mathbf{X},\mathbf{Z}\right)\right).
\end{align}
That is, for an increasing mapping $f_{\beta}\left(\cdot\right)$,
the ranking of the estimates does not change, either with the bounded
q-metric or with its associated score. As the GOSPA and T-GOSPA q-metrics
have the units of the base distance $d_{b}\left(\cdot,\cdot\right)$,
$\beta$ must have the same units such that the units are removed
for the mappings (\ref{eq:metric_preserving1})-(\ref{eq:metric_preserving4})
to be well defined. In addition, parameter $\beta$ can be set to
calibrate the similarity in the score function. For example, we can
fix $\beta$ such that an estimate with $n_{fa}$ false objects when
the ground truth has no objects has a similarity score of $s_{n_{fa}}$.
As the GOSPA q-metric error in this case is $c\sqrt[p]{\rho n_{fa}}$,
we can obtain $\beta$ by solving
\begin{align}
1-f_{\beta}\left(c\sqrt[p]{\rho n_{fa}}\right) & =s_{n_{fa}}.\label{eq:score_calibration}
\end{align}

\begin{example}
Assume that, in a given MOT system, there is a maximum localisation
error of $c=10$ $\mathrm{m}$. We use the GOSPA q-metric with $\rho=0.5$,
$p=1$. We design the similarity score with the sigmoid metric-preserving
mapping (\ref{eq:metric_preserving1}) by setting the estimate with
10 false objects, when there are no real objects, a score of 0.1.
Calibrating the score function with (\ref{eq:score_calibration})
yields $\beta=16.98\,\mathrm{m}$. 
\end{example}

\section{Q-metrics and scores for random finite sets\label{sec:Quasi-metrics-RFS}}

This section extends the GOSPA and T-GOSPA q-metrics, and their associated
scores, to RFSs of objects \cite{Mahler_book14} and trajectories
\cite{Angel20_b}. This extension is relevant for performance evaluation
via Monte Carlo simulations. It is also relevant for performance evaluation
using a dataset containing multiple scenarios, each associated with
a different ground truth. It is possible to understand these two cases
as the comparison between an RFS with the ground truth and another
RFS with the estimate \cite[Sec. V]{Angel21_f}.

\subsection{Q-metrics}

The GOSPA q-metric can be extended to RFSs of objects using the expected
GOSPA q-metric value, as done for the GOSPA metric \cite[Sec. III]{Rahmathullah17}.
We consider a real parameter $p'$ such that $1\leq p'<\infty$. The
expected value of the GOSPA q-metric to the $p'$ power is
\begin{align}
\mathbb{E}\big[d_{p}^{\left(c,\rho\right)}(\mathbf{x},\mathbf{y})^{p'}\big] & =\int\int d_{p}^{\left(c,\rho\right)}(\mathbf{x},\mathbf{y})^{p'}p\left(\mathbf{x},\mathbf{y}\right)\delta\mathbf{x}\delta\mathbf{y}\label{eq:Expected_GOSPAq}
\end{align}
where this expectation is a double set integral \cite{Mahler_book14}
taken with respect to the joint density $p\left(\mathbf{x},\mathbf{y}\right)$
of the RFSs $\mathbf{x}$ and $\mathbf{y}$. 

Similarly, we can extend the T-GOSPA q-metric to RFSs of trajectories
using the expected T-GOSPA q-metric value. The expected value of the
T-GOSPA q-metric to the $p'$ power has the same expression as (\ref{eq:Expected_GOSPAq}),
but using $\mathbf{X}$ and $\mathbf{Y}$ instead of $\mathbf{x}$
and $\mathbf{y}$, and sets integrals for sets of trajectories \cite{Angel20_b}
instead of sets integrals for sets of objects. Then, we have this
result.
\begin{lem}
\label{lem:GOSPA_TGOSPA_RFS}For a real parameter $p'$ such that
$1\leq p'<\infty$, $\left(\mathbb{E}\big[d_{p}^{\left(c,\rho\right)}(\mathbf{x},\mathbf{y})^{p'}\big]\right)^{1/p'}$
is a q-metric on the space of RFSs of objects with a finite cardinality
moment $\mathbb{E}\left[\left|\cdot\right|^{p'/p}\right]<\infty$.
In addition, $\left(\mathbb{E}\big[d_{p}^{\left(c,\rho,\gamma\right)}(\mathbf{X},\mathbf{Y})^{p'}\big]\right)^{1/p'}$
and $\left(\mathbb{E}\big[\overline{d}_{p}^{\left(c,\rho,\gamma\right)}(\mathbf{X},\mathbf{Y})^{p'}\big]\right)^{1/p'}$
are q-metrics on the space of RFSs of trajectories with a finite cardinality
moment $\mathbb{E}\left[\left|\cdot\right|^{p'/p}\right]<\infty$.
\end{lem}
The proof is equivalent to the proof of Lemma 3 in \cite{Angel20_d}. 

\subsection{Q-metric-based scores}

The extension of the q-metric-based scores to RFSs of objects and
trajectories is provided via the following lemma.
\begin{lem}
\label{lem:GOSPA_TGOSPA_score_RFS}Given the GOSPA and T-GOSPA q-metric
scores $s(\mathbf{x},\mathbf{y})=1-f_{\beta}\left(d_{p}^{\left(c,\rho\right)}(\mathbf{x},\mathbf{y})\right)$
and $s(\mathbf{X},\mathbf{Y})=1-f_{\beta}\left(d_{p}^{\left(c,\rho,\gamma\right)}\left(\mathbf{X},\mathbf{Y}\right)\right)$,
$\mathbb{E}\big[s(\mathbf{x},\mathbf{y})\big]$ and $\mathbb{E}\big[s(\mathbf{X},\mathbf{Y})\big]$
are q-metric-based scores on the space of RFSs of objects and RFSs
of trajectories, respectively.
\end{lem}
The proof is similar to the one of Lemma \ref{lem:GOSPA_TGOSPA_RFS}
considering $p'=1$. When we deal with scores and RFSs, we simply
take the expected value to keep the q-metric-based score property.
In simulation-based MOT evaluation, the expected values in Lemmas
\ref{lem:GOSPA_TGOSPA_RFS} and \ref{lem:GOSPA_TGOSPA_score_RFS}
are approximated via Monte Carlo simulation, as done in the next section.

\section{Simulations\label{sec:Simulations}}

This section examines the performance of several Bayesian MOT algorithms
via the T-GOSPA q-metric, OSPA$^{(2)}$ and UOSPA$^{(2)}$. We have
implemented the trajectory Poisson multi-Bernoulli mixture (T-PMBM)
filter \cite{Granstrom25}, the PMBM filter \cite{Williams15b,Angel18_b}
and the generalised labelled multi-Bernoulli (GLMB) filter\footnote{Code of the T-PMBM and PMBM filters can be found at https://github.com/Agarciafernandez/MTT. 

Code of the GLMB filter can be found at https://ba-tuong.vo-au.com.} \cite{Vo13}. The PMBM and GLMB both use sequential track formation
by connecting state estimates with a similar auxiliary variable (PMBM)
\cite{Angel20_e} or with the same label (GLMB). T-PMBM and PMBM consider
a maximum number of 200 global hypotheses, while GLMB has a maximum
of 1000. All filters use Murty's algorithm \cite{Murty68} in the
update step to select the global hypotheses, arising from a previous
global hypotheses, that have the highest weights. The T-PMBM implementation
uses an $L$-scan window with $L=5$, to jointly update the last $L$
time steps of single-trajectory densities. The pruning and estimation
parameters are set as in \cite{Angel20_e}. All units are in the international
system and are not included for brevity.

The state of a single object is $x=\left[p_{x},\dot{p}_{x},p_{y},\dot{p}_{y}\right]^{T}$,
containing 2-D position $\left[p_{x},p_{y}\right]^{T}$ and velocity
$\left[\dot{p}_{x},\dot{p}_{y}\right]^{T}$. Objects move with a nearly
constant velocity model \cite{Bar-Shalom_book01} with a sampling
time $\tau=1$, and the process noise intensity is $q=0.4$. The probability
of survival of the objects is 0.99. 

The birth single-object density is $p_{b}\left(x\right)=\mathcal{N}\left(x;\overline{x}_{b},P_{b}\right)$,
which represents a Gaussian density with mean $\overline{x}_{b}=\left[400,0,400,0\right]^{T}$
and covariance matrix $P_{b}=\mathrm{diag}\left(\left[300^{2},2^{2},300^{2},2^{2}\right]\right)$.
For T-PMBM and PMBM filters, the birth model is a Poisson point process
(PPP). Its intensity is $3p_{b}\left(x\right)$ at the initial time
step ($k=1$) and $0.005p_{b}\left(x\right)$ at the following time
steps. The GLMB filter uses a multi-Bernoulli birth model, which is
chosen to approximate the PPP birth model with 5 Bernoulli components
as in \cite{Angel20_e}. 

\begin{figure}
\begin{centering}
\includegraphics[scale=0.6]{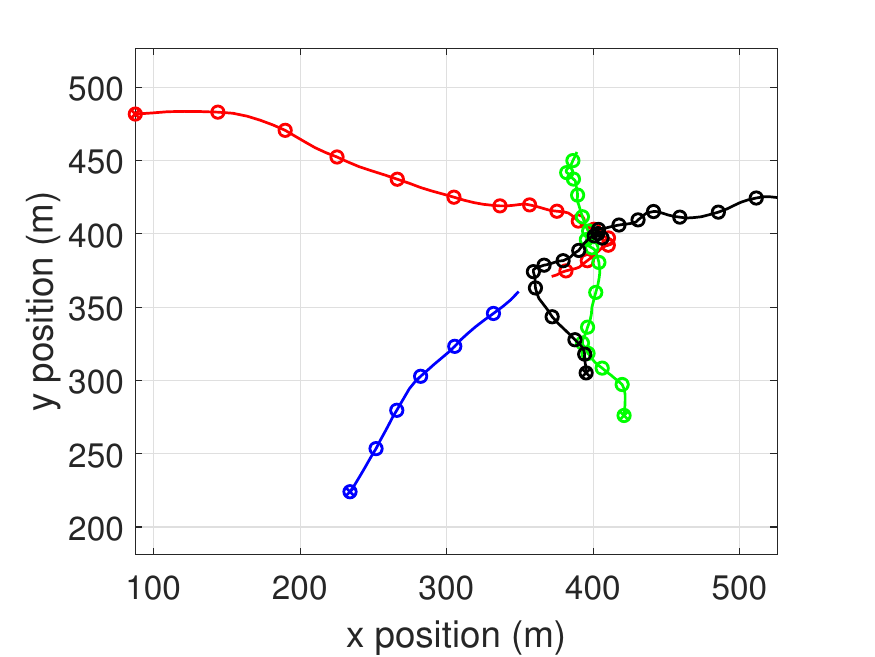}
\par\end{centering}
\caption{\label{fig:Scenario}Scenario considered the simulations. All objects
appear at time step 1 except the black one, which appears at time
step 6. The initial object states are marked with a cross. The blue
object dies at time step 30, the red at time step 75, the green at
time step 80 and the black one at time step 100.}
\end{figure}

The probability of detection of each object is set to $p^{D}=0.9$.
The sensor measures the positional elements with an additive zero-mean
Gaussian noise with covariance matrix $R=\mathrm{diag}\left(\left[4,4\right]\right)$.
Clutter follows a PPP whose intensity is $\lambda^{C}\left(z\right)=\overline{\lambda}^{C}u_{A}\left(z\right)$
where $\overline{\lambda}^{C}=20$ and $u_{A}\left(z\right)$ is a
uniform density in the area $A=\left[0,800\right]\times\left[0,800\right]$.
The simulation has 101 time steps and contains four objects. The scenario
is represented in Figure \ref{fig:Scenario}. Three of these objects
get in close proximity roughly at the middle of the simulation\footnote{Code with the scenario to reproduce the results is available at https://github.com/Agarciafernandez/MTT.}.

At each time step, we evaluate the accuracy of the estimated set $\mathbf{\hat{X}}_{k}$
of all trajectories (positional elements) up to the current time step
$k$ compared with the true set $\mathbf{X}_{k}$ of all trajectories
via Monte Carlo simulation with $N_{mc}=100$ runs. The T-GOSPA q-metric
has been implemented with the Euclidean distance, parameter $p=2$,
maximum localisation error $c=10$, track switching penalty $\gamma=1$
and q-metric parameter $\rho\in\left\{ 0.3,0.5,0.7\right\} $. We
consider three choices of $\rho$ to analyse the effect of this parameter
in the q-metric value. With $\rho=0.3$, missed objects are penalised
more than false objects. With $\rho=0.5$, missed and false objects
are penalised equally (T-GOSPA metric). With $\rho=0.7$, false objects
are penalised more than missed objects. The root mean square (RMS)
T-GOSPA q-metric at time step $k$ is
\begin{align}
d\left(k\right) & =\sqrt{\frac{1}{N_{mc}k}\sum_{i=1}^{N_{mc}}d_{2}^{\left(10,\rho\right)}\left(\mathbf{X}_{k},\mathbf{\hat{X}}_{k}^{i}\right)^{2}},\label{eq:error_time_k}
\end{align}
where the squared q-metric has been normalised by the time window
length $k$ (the time window is from time step 1 to $k$). Equation
(\ref{eq:error_time_k}) is a Monte Carlo approximation of the q-metric
for RFSs in Lemma \ref{lem:GOSPA_TGOSPA_RFS}. In this case, $\mathbf{X}_{k}$
can be considered a deterministic RFS, and $\mathbf{\hat{X}}_{k}^{i}$
are realisations of the estimated RFS of trajectories. The T-GOSPA
similarity score has been implemented with the scaled and translated
sigmoid (\ref{eq:metric_preserving1}). To account for the increasing
window length, the $\beta$ parameter of the score is chosen as $\beta=k\beta_{0}$,
where $\beta_{0}$ is chosen using (\ref{eq:score_calibration}) to
return a score of 0.1 with 50 false targets and $\rho=0.5$, resulting
in $\beta_{0}=16.98$. OSPA$^{(2)}$ and UOSPA$^{(2)}$ also use $p=2$
and $c=10$, and their RMS errors are computed similarly to (\ref{eq:error_time_k}).

\begin{figure}
\begin{centering}
\includegraphics[scale=0.6]{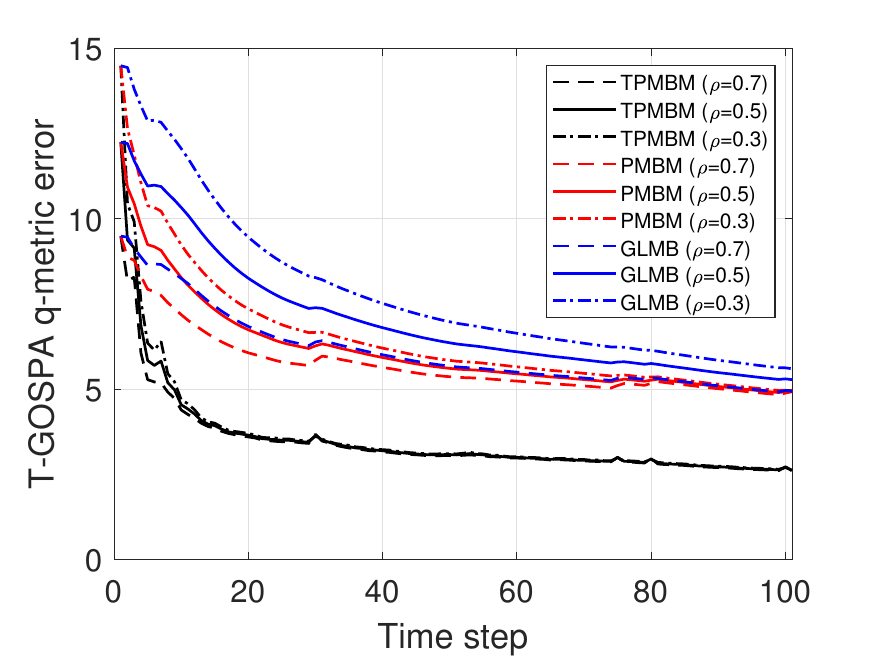}
\par\end{centering}
\begin{centering}
\includegraphics[scale=0.6]{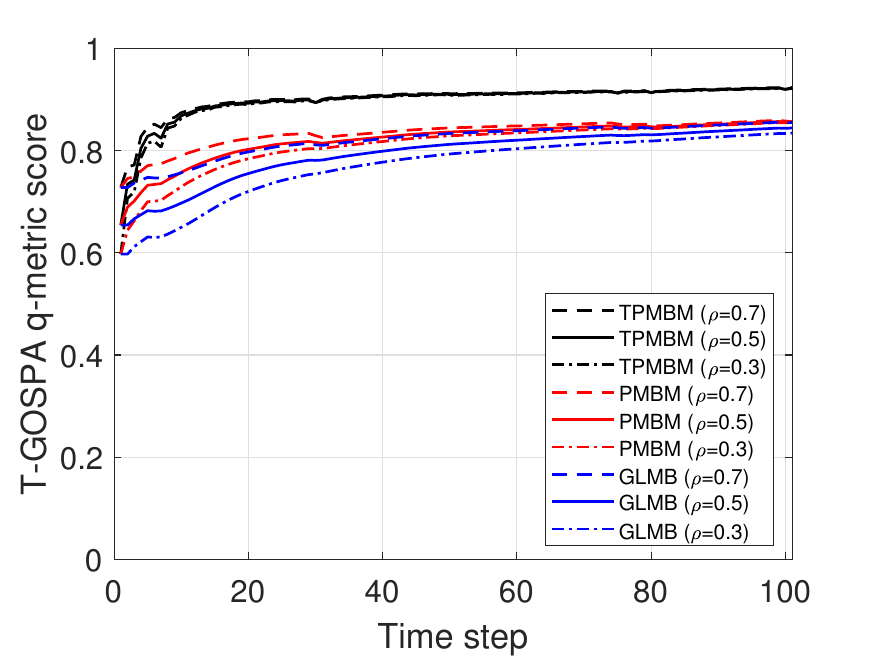}
\par\end{centering}
\caption{\label{fig:RMS-T-GOSPA-q-metric-errors}RMS-T-GOSPA q-metric errors
at each time step (top) and mean T-GOSPA q-metric scores at each time
step (bottom). The TPMBM filter performs best for all considered values
of $\rho$. }

\end{figure}

\begin{figure}
\begin{centering}
\includegraphics[scale=0.3]{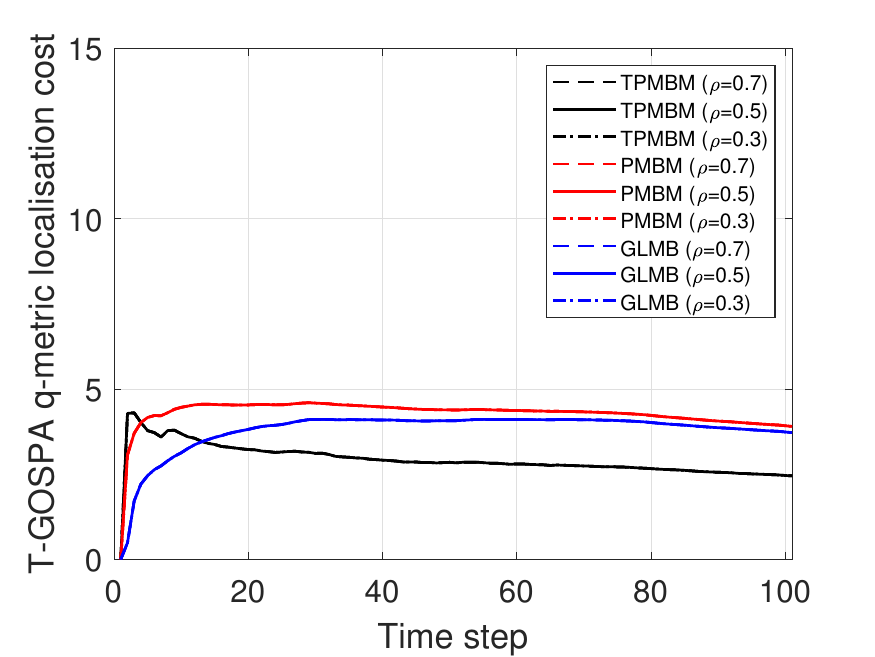}\includegraphics[scale=0.3]{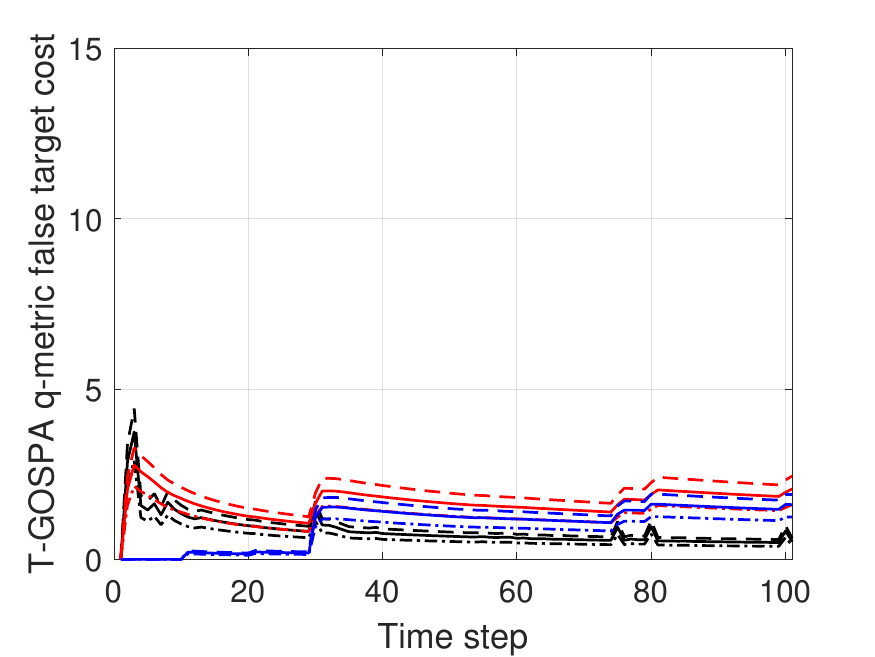}
\par\end{centering}
\begin{centering}
\includegraphics[scale=0.3]{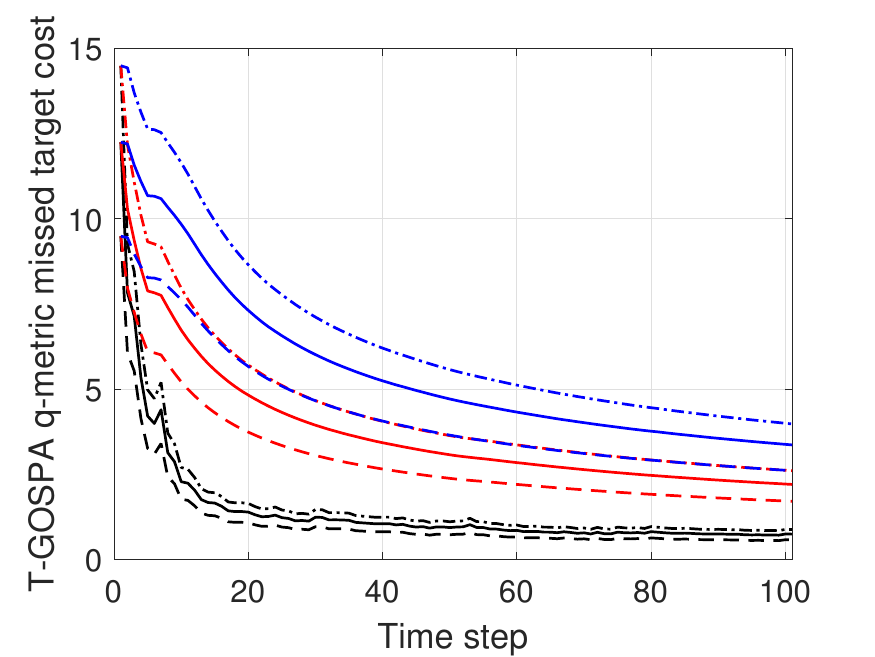}\includegraphics[scale=0.3]{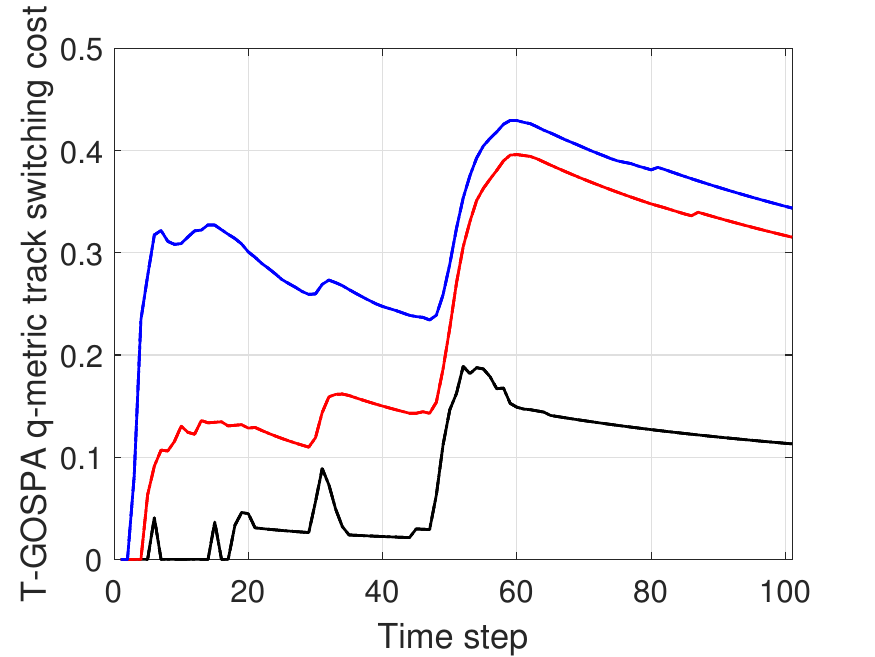}
\par\end{centering}
\caption{\label{fig:RMS-T-GOSPA-q-metric-decompositi}RMS-T-GOSPA q-metric
decomposition across time. A change in $\rho$ implies a change in
the false and missed object costs, whereas localisation and track
switching costs remain unchanged. Thus there is only one line style
visible for each filter in the first and last sub-figures.}
\end{figure}

We first analyse the effect of parameter $\rho$. The RMS-T-GOSPA
q-metric errors and the mean T-GOSPA scores at each time step are
shown in Figure \ref{fig:RMS-T-GOSPA-q-metric-errors}. We can see
that the ranking of the algorithms remains unchanged either looking
at the errors or the scores. For all choices of $\rho$, the T-PMBM
filter achieves the best performance, followed by PMBM and GLMB. The
q-metric errors are higher with $\rho=0.3$ than with the other values
of $\rho$. This means that in this scenario there are more missed
objects than false objects, which can also be checked in the error
decomposition in Figure \ref{fig:RMS-T-GOSPA-q-metric-decompositi}.
Overall, the q-metric value decreases with time since the estimation
of all the trajectories becomes more accurate (normalised by the time
window length). While localisation errors stay roughly the same across
time, the missed object cost decreases since objects are mainly missed
at the beginning of the simulation, so their normalised cost decreases
with time. At the time steps when some objects disappear, there are
some spikes in the T-GOSPA error, due to false alarms. We can see
that the spikes are larger with $\rho=0.7$ since higher $\rho$ penalises
false objects more. From Lemma \ref{lem:Optimal_assignment_TGOSPA_q},
we know that the localisation error and track switching cost do not
change with $\rho$. This can be seen in Figure \ref{fig:RMS-T-GOSPA-q-metric-decompositi}.

\begin{figure}
\begin{centering}
\includegraphics[scale=0.6]{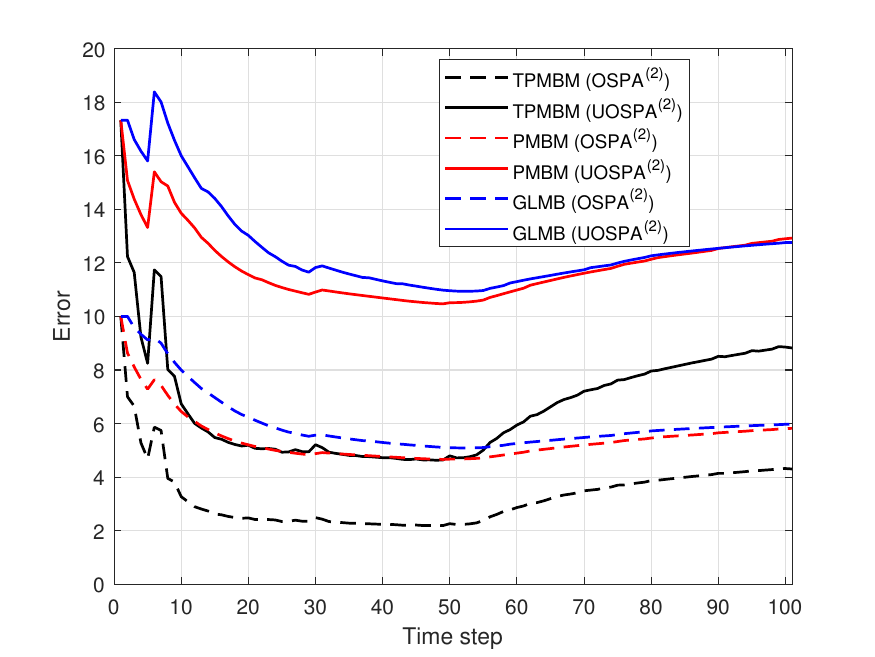}
\par\end{centering}
\caption{\label{fig:RMS-OSPA-and-RMS-UOSPA}RMS-OSPA$^{(2)}$ and RMS-UOSPA$^{(2)}$
errors at each time step. The TPMBM filter performs best.}

\end{figure}

The RMS-OSPA$^{(2)}$ and RMS-UOSPA$^{(2)}$ errors at each time step
are shown in Figure \ref{fig:RMS-OSPA-and-RMS-UOSPA}. In this case,
these metrics agree with the ranking of algorithms provided by the
T-GOSPA q-metric. While we can decompose this plot into cardinality
errors and another error for trajectory mismatch, it is not possible
to decompose the error as in Figure \ref{fig:RMS-T-GOSPA-q-metric-decompositi},
not meeting DP5.

We now proceed to evaluate tracking performance for different values
of $p^{D}$ and $\overline{\lambda}^{C}$. The RMS-GOSPA error across
all time steps is shown in Table \ref{tab:RMS-TGOSPA-q-metric-errors}.
As space allows, this table also includes the (track-oriented) PMB
filter \cite{Williams15b} with sequential track building. The best
performing filter is always the T-PMBM filter, followed by PMBM/PMB
(depending on the scenario) and then followed by GLMB. As expected,
higher clutter and lower $p^{D}$ generally increase the error (this
happens in all entries except for the T-PMBM $\rho=0.7$ and $\overline{\lambda}^{C}=20$).
In the considered scenarios, the missed object error dominates the
false object error since q-metric value is highest for $\rho=0.3$.
The RMS-OSPA$^{(2)}$ and RMS-UOSPA$^{(2)}$ errors across all time
steps are shown in Table \ref{tab:RMS-OSPA2-metric-errors}. In these
results, we can see that increasing the clutter rate does not increase
the metric value as consistently as with T-GOSPA q-metric (entries
highlighted in red). In these cases, these metrics are indicating
that the results with higher clutter rate are preferable with those
with a smaller clutter rate. A factor that is contributing to these
results for these metrics is that assignments of real to estimated
trajectories cannot change over time. All metrics and q-metrics indicate
that all algorithms with $\overline{\lambda}^{C}=15$ perform better
than the same algorithms with $\overline{\lambda}^{C}=25$, as expected.

\begin{table*}
\caption{\label{tab:RMS-TGOSPA-q-metric-errors}RMS-TGOSPA q-metric errors
across all time steps}

\begin{centering}
\par\end{centering}
\begin{centering}
\begin{tabular}{c|c|ccc|ccc|ccc|ccc}
\hline 
\multicolumn{2}{c|}{} &
\multicolumn{3}{c|}{T-PMBM $\rho=$} &
\multicolumn{3}{c|}{PMBM $\rho=$} &
\multicolumn{3}{c|}{PMB $\rho=$} &
\multicolumn{3}{c}{GLMB $\rho=$}\tabularnewline
\multicolumn{1}{c}{$p^{D}$} &
$\overline{\lambda}^{C}$ &
0.3 &
0.5 &
0.7 &
0.3 &
0.5 &
0.7 &
0.3 &
0.5 &
0.7 &
0.3 &
0.5 &
0.7\tabularnewline
\hline 
\multirow{3}{*}{0.7} &
15 &
5.69 &
5.47 &
5.25 &
8.37 &
7.71 &
7.00 &
8.35 &
7.81 &
7.23 &
9.68 &
8.61 &
7.39\tabularnewline
 & 20 &
5.72 &
5.47 &
\textcolor{red}{5.21} &
8.43 &
7.77 &
7.04 &
8.42 &
7.88 &
7.29 &
9.88 &
8.76 &
7.47\tabularnewline
 & 25 &
6.08 &
5.73 &
5.36 &
8.72 &
7.96 &
7.12 &
8.67 &
8.07 &
7.42 &
10.17 &
8.98 &
7.61\tabularnewline
\hline 
\multirow{3}{*}{0.8} &
15 &
4.57 &
4.28 &
3.98 &
7.46 &
6.89 &
6.26 &
7.43 &
6.95 &
6.43 &
8.83 &
7.85 &
6.73\tabularnewline
 & 20 &
4.79 &
4.50 &
4.18 &
7.64 &
7.03 &
6.36 &
7.61 &
7.10 &
6.55 &
9.13 &
8.09 &
6.90\tabularnewline
 & 25 &
5.26 &
4.82 &
4.34 &
7.98 &
7.29 &
6.52 &
7.83 &
7.27 &
6.66 &
9.38 &
8.29 &
7.04\tabularnewline
\hline 
\multirow{3}{*}{0.9} &
15 &
3.97 &
3.78 &
3.58 &
6.75 &
6.29 &
5.80 &
6.70 &
6.29 &
5.85 &
8.16 &
7.25 &
6.21\tabularnewline
 & 20 &
4.08 &
3.84 &
3.59 &
6.84 &
6.34 &
5.80 &
6.78 &
6.34 &
5.86 &
8.35 &
7.39 &
6.29\tabularnewline
 & 25 &
4.23 &
3.94 &
3.62 &
7.07 &
6.52 &
5.91 &
6.95 &
6.47 &
5.96 &
8.57 &
7.57 &
6.42\tabularnewline
\hline 
\multirow{3}{*}{0.95} &
15 &
3.70 &
3.51 &
3.30 &
6.40 &
5.91 &
5.37 &
6.37 &
5.92 &
5.43 &
7.83 &
6.95 &
5.94\tabularnewline
 & 20 &
3.70 &
3.51 &
3.32 &
6.43 &
5.93 &
5.38 &
6.40 &
5.95 &
5.46 &
7.98 &
7.06 &
6.00\tabularnewline
 & 25 &
3.88 &
3.66 &
3.42 &
6.62 &
6.08 &
5.52 &
6.54 &
6.07 &
5.56 &
8.18 &
7.22 &
6.12\tabularnewline
\hline 
\end{tabular}
\par\end{centering}
\centering{}
\end{table*}

\begin{table*}
\caption{\label{tab:RMS-OSPA2-metric-errors}RMS-OSPA$^{(2)}$ and RMS-UOSPA$^{(2)}$
metric errors across all time steps}

\begin{centering}
\par\end{centering}
\centering{}%
\begin{tabular}{c|c|cc|cc|cc|cc}
\hline 
\multicolumn{1}{c}{} &
 &
\multicolumn{2}{c|}{T-PMBM} &
\multicolumn{2}{c|}{PMBM} &
\multicolumn{2}{c|}{PMB} &
\multicolumn{2}{c}{GLMB}\tabularnewline
\cline{3-10}
\multicolumn{1}{c}{$p^{D}$} &
$\overline{\lambda}^{C}$ &
OSPA$^{(2)}$ &
UOSPA$^{(2)}$ &
OSPA$^{(2)}$ &
UOSPA$^{(2)}$ &
OSPA$^{(2)}$ &
UOSPA$^{(2)}$ &
OSPA$^{(2)}$ &
UOSPA$^{(2)}$\tabularnewline
\hline 
\multirow{3}{*}{0.7} &
15 &
4.20 &
8.23 &
6.28 &
13.56 &
6.41 &
14.19 &
7.23 &
15.83\tabularnewline
 & 20 &
4.35 &
8.64 &
6.52 &
14.46 &
6.58 &
14.90 &
7.29 &
15.87\tabularnewline
 & 25 &
4.44 &
8.74 &
6.64 &
14.58 &
6.77 &
15.31 &
7.43 &
16.04\tabularnewline
\hline 
\multirow{3}{*}{0.8} &
15 &
3.46 &
6.74 &
5.68 &
12.08 &
5.71 &
12.22 &
6.71 &
14.44\tabularnewline
 & 20 &
3.69 &
7.29 &
5.98 &
12.96 &
6.01 &
13.13 &
6.93 &
15.04\tabularnewline
 & 25 &
3.84 &
7.51 &
6.18 &
13.43 &
6.20 &
13.88 &
6.98 &
\textcolor{red}{14.98}\tabularnewline
\hline 
\multirow{3}{*}{0.9} &
15 &
3.23 &
6.36 &
5.45 &
11.73 &
5.45 &
11.68 &
6.16 &
12.97\tabularnewline
 & 20 &
3.51 &
7.04 &
5.51 &
11.95 &
5.45 &
11.90 &
\textcolor{red}{6.14} &
\textcolor{red}{12.69}\tabularnewline
 & 25 &
\textcolor{red}{3.26} &
\textcolor{red}{6.42} &
5.54 &
\textcolor{red}{11.85} &
5.70 &
12.48 &
6.31 &
13.18\tabularnewline
\hline 
\multirow{3}{*}{0.95} &
15 &
3.00 &
5.87 &
4.77 &
9.73 &
4.88 &
10.01 &
5.97 &
12.42\tabularnewline
 & 20 &
3.09 &
6.17 &
5.01 &
10.57 &
5.17 &
10.98 &
\textcolor{red}{5.96} &
\textcolor{red}{12.40}\tabularnewline
 & 25 &
3.13 &
\textcolor{red}{6.14} &
5.30 &
11.31 &
5.32 &
11.40 &
6.10 &
12.65\tabularnewline
\hline 
\end{tabular}
\end{table*}

\section{Conclusions\label{sec:Conclusions}}

This paper has presented two quasi-metrics for performance evaluation
of MOT algorithms, with the important characteristic of being clearly
interpretable to penalise localisation errors, the number of missed
and false objects, and track switches. This interpretability enables
a fair comparison of the estimated sets of objects or sets of trajectories
provided by different algorithms since these q-metrics promote algorithms
with lower values of the indicated errors. The proposed quasi-metrics
are extensions of the GOSPA metrics for sets of objects, and sets
of trajectories, and allow uneven costs for the missed and false objects.
The paper proves the identity and triangle inequality properties that
are required to define quasi-metrics. The paper has also presented
how to define similarity score functions based on the GOSPA and T-GOSPA
quasi-metrics. The proposed quasi-metrics and their scores have also
been extended to RFSs of objects and RFSs of trajectories.

The T-GOSPA q-metric has been applied to evaluate MOT simulation results
for different values of $\rho$, showing the effects on the q-metric
value of using different costs for missed and false objects. The GOSPA
and T-GOSPA quasi-metrics should be used in applications in which
the user is interested in penalising missed and false objects differently,
according to design principles DP1-DP5. This can be achieved by choosing
the parameter $\rho$ accordingly. In other cases, our recommendation
for evaluation of MOT algorithms is to use the GOSPA and T-GOSPA metrics
as well as their decompositions into their different components to
provide a more thorough analysis. It is also true that there can be
applications in which users prefer metrics with different design principles
from DP1-DP5, for instance, metrics that penalise cardinality mismatch
instead of the number of missed and false objects. In these cases,
it can be reasonable to use other metric, for instance, OSPA, UOSPA,
COLA or GOSPA ($\alpha\neq2$).

A direction of future work is to extend the proposed quasi-metrics
to uncertainty-aware MOT performance evaluation, as done for GOSPA
and T-GOSPA in \cite{Xia25,Xia25b}. Another direction of future work
is the application of these quasi-metrics to computer vision MOT performance
evaluation. In addition, future work can develop of other quasi-metrics
for MOT performance evaluation, instead of the standard (symmetric)
metrics.

\bibliographystyle{IEEEtran}
\bibliography{11C__Trabajo_laptop_Mis_articulos_Finished_TGOSPA_q-metric_Accepted_Referencias}

\cleardoublepage{}

{\LARGE Supplemental material: ``GOSPA and T-GOSPA quasi-metrics for
evaluation of multi-object tracking algorithms''}{\LARGE\par}

\appendices{}

\section{\label{sec:AppendixA}}

In the proof of the triangle inequality, we use the definition of
the GOSPA q-metric in terms of permutations instead of assignment
sets. In this appendix, we introduce this alternative expression.
We first introduce the cut-off base q-metric $d_{b}^{\left(c\right)}\left(x,y\right)=\min\left(d_{b}(x,y),c\right)$.
The set of all permutations on $\mathbb{N}_{n}=\left\{ 1,...,n\right\} $
is denoted by $\prod_{n}$ such that a permutation is written as $\pi=\left(\pi(1),...,\pi(n)\right)\in\prod_{n}$,
with $\pi(i)\in\mathbb{N}_{n}$, $i\in\mathbb{N}_{n}$. 
\begin{lem}
\label{lem:GOSPA_q_permutation} For $\left|\mathbf{y}\right|\geq\left|\mathbf{x}\right|$,
the GOSPA q-metric in (\ref{eq:GOSPA_quasimetric}) in permutation
form is
\begin{align}
 & d_{p}^{\left(c,\rho\right)}\left(\mathbf{x},\mathbf{y}\right)\nonumber \\
 & =\min_{\pi\in\prod_{\left|\mathbf{y}\right|}}\left(\sum_{i=1}^{\left|\mathbf{x}\right|}d_{b}^{\left(c\right)}\left(x_{i},y_{\pi\left(i\right)}\right)^{p}+\rho c^{p}\left(\left|\mathbf{y}\right|-\left|\mathbf{x}\right|\right)\right)^{1/p}\label{eq:GOSPA_quasimetric_permutation1}
\end{align}
and, for $\left|\mathbf{y}\right|\leq\left|\mathbf{x}\right|$, the
GOSPA q-metric is
\begin{align}
 & d_{p}^{\left(c,\rho\right)}\left(\mathbf{x},\mathbf{y}\right)\nonumber \\
 & =\min_{\pi\in\prod_{\left|\mathbf{x}\right|}}\left(\sum_{i=1}^{\left|\mathbf{y}\right|}d_{b}^{\left(c\right)}\left(y_{i},x_{\pi\left(i\right)}\right)^{p}+\left(1-\rho\right)c^{p}\left(\left|\mathbf{x}\right|-\left|\mathbf{y}\right|\right)\right)^{1/p}.
\end{align}
\end{lem}
The proof of this lemma follows the proof of how to write the GOSPA
metric as an optimisation over assignment sets instead of permutations
\cite[App. B]{Rahmathullah17}.

\section{\label{sec:AppendixB}}

In this appendix, we prove the triangle inequality for the GOSPA q-metric,
using its expression in terms of permutations, see (\ref{eq:GOSPA_quasimetric_permutation1}).
In particular we prove that for any sets $\mathbf{x}$, $\mathbf{y}$
and $\mathbf{z}$, the following inequality holds
\begin{align}
d_{p}^{\left(c,\rho\right)}\left(\mathbf{x},\mathbf{y}\right) & \leq d_{p}^{\left(c,\rho\right)}\left(\mathbf{x},\mathbf{z}\right)+d_{p}^{\left(c,\rho\right)}\left(\mathbf{z},\mathbf{y}\right).\label{eq:triangle_inequality_append}
\end{align}
We proceed as in the proof of triangle inequality for the GOSPA metric
\cite[App. A]{Rahmathullah17}. 

The Minkowski\textquoteright s inequality for two sequences of different
lengths $\left(a_{1},...,a_{m}\right)$ and $\left(b_{1},...,b_{n}\right)$
with $m\leq n$, and $1\leq p\leq\infty$ is required in the proof
and is given by \cite{Kubrusly_book11}
\begin{align}
 & \left(\sum_{i=1}^{m}|a_{i}+b_{i}|^{p}+\sum_{i=m+1}^{n}|b_{i}|^{p}\right)^{1/p}\nonumber \\
 & \leq\left(\sum_{i=1}^{m}|a_{i}|^{p}+\sum_{i=1}^{n}|b_{i}|^{p}\right)^{1/p}.\label{eq:Minkowski}
\end{align}
We first assume that $|\mathbf{y}|\geq|\mathbf{x}|$ and consider
three different cases depending on the cardinality of $\mathbf{z}$
in relation to the cardinality of $\mathbf{x}$ and $\mathbf{y}$. 

\subsection{Case 1: $|\mathbf{x}|\protect\leq|\mathbf{y}|\protect\leq|\mathbf{z}|$}

For any $\pi\in\Pi_{|\mathbf{y}|}$, we have
\begin{align}
 & d_{p}^{\left(c,\rho\right)}\left(\mathbf{x},\mathbf{y}\right)\nonumber \\
 & \leq\left(\sum_{i=1}^{\left|\mathbf{x}\right|}d_{b}^{\left(c\right)}\left(x_{i},y_{\pi\left(i\right)}\right)^{p}+\rho c^{p}\left(\left|\mathbf{y}\right|-\left|\mathbf{x}\right|\right)\right)^{1/p}.
\end{align}
Using the triangle inequality on the single-object q-metric, and any
permutation $\sigma\in\Pi_{|\mathbf{z}|}$, we obtain
\begin{align}
 & d_{p}^{\left(c,\rho\right)}\left(\mathbf{x},\mathbf{y}\right)\nonumber \\
 & \leq\left(\sum_{i=1}^{\left|\mathbf{x}\right|}\left[d_{b}^{\left(c\right)}\left(x_{i},z_{\sigma\left(i\right)}\right)+d_{b}^{\left(c\right)}\left(z_{\sigma\left(i\right)},y_{\pi\left(i\right)}\right)\right]^{p}\right.\nonumber \\
 & \left.\quad+\rho c^{p}\left(\left|\mathbf{y}\right|-\left|\mathbf{x}\right|\right)\vphantom{\sum_{i=1}^{\left|\mathbf{x}\right|}}\right)^{1/p}\nonumber \\
 & \leq\left(\sum_{i=1}^{\left|\mathbf{x}\right|}\left[d_{b}^{\left(c\right)}\left(x_{i},z_{\sigma\left(i\right)}\right)+d_{b}^{\left(c\right)}\left(z_{\sigma\left(i\right)},y_{\pi\left(i\right)}\right)\right]^{p}\right.\nonumber \\
 & \left.\quad+\rho c^{p}\left(\left|\mathbf{y}\right|-\left|\mathbf{x}\right|\right)+c^{p}\left(\left|\mathbf{z}\right|-\left|\mathbf{y}\right|\right)\vphantom{\sum_{i=1}^{\left|X\right|}}\right)^{1/p}\nonumber \\
 & =\left(\sum_{i=1}^{\left|\mathbf{x}\right|}\left[d_{b}^{\left(c\right)}\left(x_{i},z_{\sigma\left(i\right)}\right)+d_{b}^{\left(c\right)}\left(z_{\pi^{-1}\left(\sigma\left(i\right)\right)},y_{i}\right)\right]^{p}\right.\nonumber \\
 & \left.\quad+\rho c^{p}\left(\left|\mathbf{z}\right|-\left|\mathbf{x}\right|\right)+\left(1-\rho\right)c^{p}\left(\left|\mathbf{z}\right|-\left|\mathbf{y}\right|\right)\vphantom{\sum_{i=1}^{\left|X\right|}}\right)^{1/p}
\end{align}
where $\pi^{-1}$ is the inverse permutation of $\pi$. Now, we define
the permutation $\tau\in\Pi_{|\mathbf{z}|}$ such that the first $|\mathbf{x}|$
entries are $\pi^{-1}\left(\sigma\left(i\right)\right)$ for $i=1,...,|\mathbf{x}|$.
Then, we can write
\begin{align}
 & d_{p}^{\left(c,\rho\right)}\left(\mathbf{x},\mathbf{y}\right)\nonumber \\
 & \leq\left(\sum_{i=1}^{\left|\mathbf{x}\right|}\left[d_{b}^{\left(c\right)}\left(x_{i},z_{\sigma\left(i\right)}\right)+d_{b}^{\left(c\right)}\left(z_{\tau(i)},y_{i}\right)\right]^{p}\right.\nonumber \\
 & \quad+\rho c^{p}\left(\left|\mathbf{z}\right|-\left|\mathbf{x}\right|\right)+\left(1-\rho\right)c^{p}\left(\left|\mathbf{z}\right|-\left|\mathbf{y}\right|\right)\nonumber \\
 & \quad\left.+\sum_{i=|\mathbf{x}|+1}^{|\mathbf{y}|}d_{b}^{\left(c\right)}\left(z_{\tau\left(i\right)},y_{i}\right)^{p}\right)^{1/p}\nonumber \\
 & \leq\left(\sum_{i=1}^{\left|\mathbf{x}\right|}d_{b}^{\left(c\right)}\left(x_{i},z_{\sigma\left(i\right)}\right)+\rho c^{p}\left(\left|\mathbf{z}\right|-\left|\mathbf{x}\right|\right)\right)^{1/p}\nonumber \\
 & +\left(\sum_{i=1}^{\left|\mathbf{y}\right|}d_{b}^{\left(c\right)}\left(z_{\tau\left(i\right)},y_{i}\right)+\left(1-\rho\right)c^{p}\left(\left|\mathbf{z}\right|-\left|\mathbf{y}\right|\right)\right)^{1/p}
\end{align}
where the last inequality has been obtained using (\ref{eq:Minkowski}). 

This inequality holds for any permutation $\sigma$ and $\tau$, including
the ones that minimise the first term and the second term respectively.
Therefore, this shows (\ref{eq:triangle_inequality_append}) for the
considered case.

\subsection{Case 2: $|\mathbf{x}|\protect\leq|\mathbf{z}|\protect\leq|\mathbf{y}|$}

We proceed as in Case 1. For any $\pi\in\Pi_{|\mathbf{y}|}$ and $\sigma\in\Pi_{|\mathbf{z}|}$,
we have
\begin{align}
 & d_{p}^{\left(c,\rho\right)}\left(\mathbf{x},\mathbf{y}\right)\nonumber \\
 & \leq\left(\sum_{i=1}^{\left|\mathbf{x}\right|}\left[d_{b}^{\left(c\right)}\left(x_{i},z_{\sigma\left(i\right)}\right)+d_{b}^{\left(c\right)}\left(z_{\sigma\left(i\right)},y_{\pi\left(i\right)}\right)\right]^{p}\right.\nonumber \\
 & \text{\ensuremath{\quad}}\left.+\rho c^{p}\left(\left|\mathbf{y}\right|-\left|\mathbf{x}\right|\right)\vphantom{\sum_{i=1}^{\left|X\right|}}\right)^{1/p}\nonumber \\
 & =\left(\sum_{i=1}^{\left|\mathbf{x}\right|}\left[d_{b}^{\left(c\right)}\left(x_{i},z_{\sigma\left(i\right)}\right)+d_{b}^{\left(c\right)}\left(z_{\sigma\left(i\right)},y_{\pi\left(i\right)}\right)\right]^{p}\right.\nonumber \\
 & \text{\ensuremath{\quad}}\left.+\rho c^{p}\left(\left|\mathbf{z}\right|-\left|\mathbf{x}\right|\right)+\rho c^{p}\left(|\mathbf{y}|-|\mathbf{z}|\right)\right)^{1/p}\nonumber \\
 & \leq\left(\sum_{i=1}^{\left|\mathbf{x}\right|}\left[d_{b}^{\left(c\right)}\left(x_{i},z_{\sigma\left(i\right)}\right)+d_{b}^{\left(c\right)}\left(z_{i},y_{\tau\left(i\right)}\right)\right]^{p}\right.\nonumber \\
 & +\sum_{i=|\mathbf{x}|+1}^{|\mathbf{z}|}d_{b}^{\left(c\right)}\left(z_{i},y_{\tau\left(i\right)}\right)^{p}+\rho c^{p}\left(\left|\mathbf{z}\right|-\left|\mathbf{x}\right|\right)\nonumber \\
 & \text{\ensuremath{\quad}}\left.+\rho c^{p}\left(|\mathbf{y}|-|\mathbf{z}|\right)\right)^{1/p}\nonumber \\
 & \leq\left(\sum_{i=1}^{\left|\mathbf{x}\right|}d_{b}^{\left(c\right)}\left(x_{i},z_{\sigma\left(i\right)}\right)+\rho c^{p}\left(\left|\mathbf{z}\right|-\left|\mathbf{x}\right|\right)\right)^{1/p}\nonumber \\
 & \text{\ensuremath{\quad}}+\left(\sum_{i=1}^{\left|\mathbf{z}\right|}d_{b}^{\left(c\right)}\left(z_{i},y_{\tau\left(i\right)}\right)+\rho c^{p}\left(|\mathbf{y}|-|\mathbf{z}|\right)\right)^{1/p}
\end{align}
where $\tau\in\Pi_{|\mathbf{y}|}$ is a permutation such that the
first $|\mathbf{x}|$ entries are $\sigma^{-1}\left(\pi\left(i\right)\right)$
for $i=1,...,|\mathbf{x}|$. As in the previous case, this holds for
any permutation $\sigma$ and $\tau$, which shows (\ref{eq:triangle_inequality_append})
for the considered case.

\subsection{Case 3: $|\mathbf{z}|\protect\leq|\mathbf{x}|\protect\leq|\mathbf{y}|$}

We now obtain
\begin{align}
 & d_{p}^{\left(c,\rho\right)}\left(\mathbf{x},\mathbf{y}\right)\nonumber \\
 & \leq\left(\sum_{i=1}^{\left|\mathbf{x}\right|}d_{b}^{\left(c\right)}\left(x_{i},y_{\pi\left(i\right)}\right)^{p}+\rho c^{p}\left(\left|\mathbf{y}\right|-\left|\mathbf{x}\right|\right)\right)^{1/p}\nonumber \\
 & \leq\left(\sum_{i=1}^{\left|\mathbf{z}\right|}d_{b}^{\left(c\right)}\left(x_{i},y_{\pi\left(i\right)}\right)^{p}+c^{p}\left(|\mathbf{x}|-|\mathbf{z}|\right)\right.\nonumber \\
 & \quad\left.+\rho c^{p}\left(\left|\mathbf{y}\right|-\left|\mathbf{x}\right|\right)\vphantom{\sum_{i=1}^{\left|Z\right|}}\right)^{1/p}
\end{align}
where we have used that $d_{b}^{\left(c\right)}\left(x_{i},y_{\pi\left(i\right)}\right)^{p}\leq c^{p}$
for $i=|\mathbf{z}|+1,...,|\mathbf{x}|$. Then we obtain
\begin{align}
 & d_{p}^{\left(c,\rho\right)}\left(\mathbf{x},\mathbf{y}\right)\nonumber \\
 & \leq\left(\sum_{i=1}^{\left|\mathbf{z}\right|}d_{b}^{\left(c\right)}\left(x_{i},y_{\pi\left(i\right)}\right)^{p}+\left(1-\rho\right)c^{p}\left(|\mathbf{x}|-|\mathbf{z}|\right)\right.\nonumber \\
 & \quad\left.+\rho c^{p}\left(|\mathbf{y}|-|\mathbf{z}|\right)\vphantom{\sum_{i=1}^{\left|\mathbf{z}\right|}}\right)^{1/p}\nonumber \\
 & \leq\left(\sum_{i=1}^{\left|\mathbf{z}\right|}\left[d_{b}^{\left(c\right)}\left(x_{i},z_{\sigma(i)}\right)+d^{\left(c\right)}\left(z_{\sigma(i)},y_{\pi\left(i\right)}\right)\right]^{p}\right.\nonumber \\
 & \quad+\left(1-\rho\right)c^{p}\left(|\mathbf{x}|-|\mathbf{z}|\right)+\rho c^{p}\left(|\mathbf{y}|-|\mathbf{z}|\right)\left.\vphantom{\sum_{i=1}^{\left|\mathbf{z}\right|}}\right)^{1/p}.
\end{align}

Proceeding as in the previous case, we prove the triangle inequality
for this case.

\subsection{Case 4: $|\mathbf{y}|\protect\leq|\mathbf{x}|$}

The previous three cases show the triangle inequality for $|\mathbf{x}|\leq|\mathbf{y}|$
for any $\rho\in\left(0,1\right)$. If $|\mathbf{y}|\leq|\mathbf{x}|$
and using Lemma (\ref{lem:GOSPA_q_symmetry}), we have that 
\begin{align}
d_{p}^{\left(c,\rho\right)}\left(\mathbf{x},\mathbf{y}\right) & =d_{p}^{\left(c,1-\rho\right)}\left(\mathbf{y},\mathbf{x}\right)\nonumber \\
 & \leq d_{p}^{\left(c,1-\rho\right)}\left(\mathbf{y},\mathbf{z}\right)+d_{p}^{\left(c,1-\rho\right)}\left(\mathbf{z},\mathbf{x}\right)\nonumber \\
 & =d_{p}^{\left(c,\rho\right)}\left(\mathbf{z},\mathbf{y}\right)+d_{p}^{\left(c,\rho\right)}\left(\mathbf{x},\mathbf{z}\right)
\end{align}
where is the second step we have used the previous derivation of the
triangle inequality. This finishes the proof of the triangle inequality
of the GOSPA q-metric.

\section{\label{sec:AppendixC}}

This appendix contains the proofs of Lemmas \ref{lem:Optimal_assignment_GOSPA_q}
and \ref{lem:Symmetrisation_GOSPA} related to the GOSPA q-metric. 

\subsection{Proof of Lemma \ref{lem:Optimal_assignment_GOSPA_q}\label{subsec:AppendixB_2}}

In this appendix, we prove Lemma \ref{lem:Optimal_assignment_GOSPA_q}.
We consider the case $\left|\mathbf{y}\right|\geq\left|\mathbf{x}\right|$
and, using (\ref{eq:GOSPA_quasimetric_permutation1}), we write the
optimal permutation as

\begin{align}
\pi^{*} & =\underset{\pi\in\prod_{\left|\mathbf{y}\right|}}{\arg\min}\left(\sum_{i=1}^{\left|\mathbf{x}\right|}d_{b}^{\left(c\right)}\left(x_{i},y_{\pi\left(i\right)}\right)^{p}+\rho c^{p}\left(\left|\mathbf{y}\right|-\left|\mathbf{x}\right|\right)\right)\\
 & =\underset{\pi\in\prod_{\left|\mathbf{y}\right|}}{\arg\min}\left(\sum_{i=1}^{\left|\mathbf{x}\right|}d_{b}^{\left(c\right)}\left(x_{i},y_{\pi\left(i\right)}\right)^{p}\right)
\end{align}
where we removed the second term as it does not depend on $\pi$.
Thus, the optimal permutation $\pi^{*}$ and consequently the optimal
assignment do not depend on $\rho$. The proof for the case $\left|\mathbf{y}\right|\leq\left|\mathbf{x}\right|$
is analogous.

\subsection{Proof of Lemma \ref{lem:Symmetrisation_GOSPA}\label{subsec:Appendix_B_3}}

In this appendix, we prove Lemma \ref{lem:Symmetrisation_GOSPA}.
We provide the proof for $\left|\mathbf{y}\right|\geq\left|\mathbf{x}\right|$,
as the proof for $\left|\mathbf{y}\right|<\left|\mathbf{x}\right|$
follows similar steps. From (\ref{eq:GOSPA_quasimetric_permutation1}),
the GOSPA q-metric to the $p$-th power is
\begin{align}
d_{p}^{\left(c,\rho\right)}\left(\mathbf{x},\mathbf{y}\right)^{p} & =\rho c^{p}\left(\left|\mathbf{y}\right|-\left|\mathbf{x}\right|\right)+\min_{\pi\in\prod_{\left|\mathbf{y}\right|}}\sum_{i=1}^{\left|\mathbf{x}\right|}d_{b}^{\left(c\right)}\left(x_{i},y_{\pi\left(i\right)}\right)^{p}.
\end{align}
In addition, using Lemma \ref{lem:GOSPA_q_symmetry}, $d_{p}^{\left(c,\rho\right)}\left(\mathbf{y},\mathbf{x}\right)=d_{p}^{\left(c,1-\rho\right)}\left(\mathbf{x},\mathbf{y}\right)$.
Therefore,
\begin{align}
d_{p}^{\left(c,\rho\right)}\left(\mathbf{y},\mathbf{x}\right)^{p} & =\left(1-\rho\right)c^{p}\left(\left|\mathbf{y}\right|-\left|\mathbf{x}\right|\right)\nonumber \\
 & \quad+\min_{\pi\in\prod_{\left|\mathbf{y}\right|}}\sum_{i=1}^{\left|\mathbf{x}\right|}d_{b}^{\left(c\right)}\left(x_{i},y_{\pi\left(i\right)}\right)^{p}.
\end{align}

Substituting the above two equations in (\ref{eq:symmetrisation_GOSPAq}),
we obtain the GOSPA metric $d_{p}^{\left(c,1/2\right)}\left(\mathbf{x},\mathbf{y}\right)$
proving Lemma \ref{lem:Optimal_assignment_GOSPA_q}.

\section{\label{sec:AppendixD}}

This appendix sketches the proof of the triangle inequality for the
T-GOSPA q-metric in LP form (\ref{eq:LP_quasimetric}). The proof
is analogous for the T-GOSPA metric \cite[App. B]{Angel20_d} so we
mainly highlight the differences here. The proof for the multi-dimensional
assignment form is similar \cite[App. B]{Angel20_d}.

Let $\overline{d}_{p}^{\left(c,\rho,\gamma\right)}\left(\mathbf{X},\mathbf{Y},W^{1:T}\right)$
denote the T-GOSPA q-metric function in (\ref{eq:LP_quasimetric})
as a function of the sequence of matrices $W^{1:T}=\left(W^{1},...,W^{T}\right)$.
Let $W_{\mathbf{X},\mathbf{Z}}^{\star,k}$ and $W_{\mathbf{Z},\mathbf{Y}}^{\star,k}$
be the matrices that minimise $\overline{d}_{p}^{\left(c,\rho,\gamma\right)}\left(\mathbf{Y},\mathbf{Z},W^{1:T}\right)$
and $\overline{d}_{p}^{\left(c,\rho,\gamma\right)}\left(\mathbf{Z},\mathbf{Y},W^{1:T}\right)$.
Using the matrices $W_{\mathbf{X},\mathbf{Z}}^{\star,k}$ and $W_{\mathbf{Z},\mathbf{Y}}^{\star,k}$,
we build a matrix $W_{\mathbf{X},\mathbf{Y}}^{k}$ that meets
\begin{align}
W_{\mathbf{X},\mathbf{Y}}^{k}(i,j) & =\sum_{l=1}^{n_{\mathbf{Z}}}W_{\mathbf{X},\mathbf{Z}}^{\star,k}(i,l)W_{\mathbf{Z},\mathbf{Y}}^{\star,k}(l,j)\label{eq:W_XY_append1}
\end{align}
for $i\in\{1,...,n_{\mathbf{X}}\}$ and $j\in\{1,...,n_{\mathbf{Y}}\}$.
The other entries of $W_{\mathbf{X},\mathbf{Y}}^{k}$ are
\begin{align}
 & W_{\mathbf{X},\mathbf{Y}}^{k}(i,j)\nonumber \\
 & =\begin{cases}
1-\sum_{j=1}^{n_{\mathbf{Y}}}W_{\mathbf{X},\mathbf{Y}}^{k}(i,j) & i\in\{1,...,n_{\mathbf{X}}\},j=n_{\mathbf{Y}}+1\\
1-\sum_{i=1}^{n_{\mathbf{X}}}W_{\mathbf{X},\mathbf{Y}}^{k}(i,j) & i=n_{\mathbf{X}}+1,j\in\{1,...,n_{\mathbf{Y}}\}\\
0 & i=n_{\mathbf{X}}+1,j=n_{\mathbf{Y}}+1.
\end{cases}\label{eq:W_XY_append2}
\end{align}
Then, it can be shown that
\begin{align}
\overline{d}_{p}^{\left(c,\rho,\gamma\right)}\left(\mathbf{X},\mathbf{Y},W_{\mathbf{X},\mathbf{Y}}^{1:T}\right) & \leq\overline{d}_{p}^{\left(c,\rho,\gamma\right)}\left(\mathbf{X},\mathbf{Z}\right)+\overline{d}_{p}^{\left(c,\rho,\gamma\right)}\left(\mathbf{Z},\mathbf{Y}\right).\label{eq:triIneq1}
\end{align}
Proving this result directly shows the triangle inequality as, by
definition, $\overline{d}_{p}^{\left(c,\rho,\gamma\right)}\left(\mathbf{X},\mathbf{Y}\right)\leq\overline{d}_{p}^{\left(c,\rho,\gamma\right)}\left(\mathbf{X},\mathbf{Y},W_{\mathbf{X},\mathbf{Y}}^{1:T}\right)$.
The proof requires the use of two inequalities, one for the localisation
cost, and another one for the track switching cost, which are then
combined. Then, one applies the Minkowski inequality to the result.
The track switching inequality in \cite[Eq. (33)]{Angel20_d} holds,
as the switching cost is the same in the metric and q-metric. The
combination of both inequalities using the Minkowski inequality is
also the same. The localisation cost inequality \cite[Eq. (46)]{Angel20_d}
only requires the following conditions: triangle inequality of the
base q-metric, $D_{\mathbf{X},\mathbf{Y}}^{k}\left(n_{\mathbf{X}}+1,j\right)=D_{\mathbf{Z},\mathbf{Y}}^{k}\left(n_{\mathbf{Z}}+1,j\right)$
and $D_{\mathbf{X},\mathbf{Y}}^{k}\left(i,n_{\mathbf{Y}}+1\right)=D_{\mathbf{X},\mathbf{Z}}^{k}\left(i,n_{\mathbf{Z}}+1\right)$.
All these properties are also met in the T-GOSPA q-metric, which implies
that (\ref{eq:triIneq1}) holds, and the triangle inequality holds.

If we used time-weights in the definition of the T-GOSPA q-metric,
the proof would be analogous to the one in \cite{Angel21_f}.

\section{\label{sec:AppendixE}}

In this appendix, we prove Lemmas \ref{lem:Optimal_assignment_TGOSPA_q}
and \ref{lem:Symmetrisation_TGOSPA} related to the T-GOSPA q-metric.

\subsection{Proof of Lemma \ref{lem:Optimal_assignment_TGOSPA_q}\label{subsec:Appendix_D1}}

In this appendix, we prove Lemma \ref{lem:Optimal_assignment_TGOSPA_q}.
In the cost function in (\ref{eq:LP_quasimetric}), the only term
that depends on $\rho$ is $\mathrm{tr}\big[\big(D_{\mathbf{X},\mathbf{Y}}^{k}\big)^{\dagger}W^{k}\big]$
that can be written as
\begin{align}
\mathrm{tr}\big[\big(D_{\mathbf{X},\mathbf{Y}}^{k}\big)^{\dagger}W^{k}\big] & =\sum_{i=1}^{n_{\mathbf{X}}+1}\sum_{j=1}^{n_{\mathbf{Y}}+1}W^{k}(i,j)D_{\mathbf{X},\mathbf{Y}}^{k}(i,j)\nonumber \\
 & =\sum_{i=1}^{n_{\mathbf{X}}}\sum_{j=1}^{n_{\mathbf{Y}}}W^{k}(i,j)D_{\mathbf{X},\mathbf{Y}}^{k}(i,j)\nonumber \\
 & +\sum_{j=1}^{n_{\mathbf{Y}}}W^{k}(n_{\mathbf{X}}+1,j)D_{\mathbf{X},\mathbf{Y}}^{k}(n_{\mathbf{X}}+1,j)\nonumber \\
 & +\sum_{i=1}^{n_{\mathbf{X}}}W^{k}(i,n_{\mathbf{Y}}+1)D_{\mathbf{X},\mathbf{Y}}^{k}(i,n_{\mathbf{Y}}+1).
\end{align}
We now expand these sums over the cases where $\mathbf{x}_{i}^{k}\neq\emptyset$,
$\mathbf{x}_{i}^{k}=\emptyset$, $\mathbf{y}_{j}^{k}\neq\emptyset$
and $\mathbf{y}_{j}^{k}=\emptyset$. In addition, we use the corresponding
value of $D_{\mathbf{X},\mathbf{Y}}^{k}(i,j)$, see (\ref{eq:D_i_j}),
when at least one of these sets is empty. This yields
\begin{align}
\mathrm{tr}\big[\big(D_{\mathbf{X},\mathbf{Y}}^{k}\big)^{\dagger}W^{k}\big] & =\sum_{i=1:\mathbf{x}_{i}^{k}\neq\emptyset}^{n_{\mathbf{X}}}\sum_{j=1:\mathbf{y}_{j}^{k}\neq\emptyset}^{n_{\mathbf{Y}}}W^{k}(i,j)D_{\mathbf{X},\mathbf{Y}}^{k}(i,j)\nonumber \\
 & +\rho c^{p}\sum_{i=1:\mathbf{x}_{i}^{k}=\emptyset}^{n_{\mathbf{X}}}\sum_{j=1:\mathbf{y}_{j}^{k}\neq\emptyset}^{n_{\mathbf{Y}}}W^{k}(i,j)\nonumber \\
 & +\left(1-\rho\right)c^{p}\sum_{i=1:\mathbf{x}_{i}^{k}\neq\emptyset}^{n_{\mathbf{X}}}\sum_{j=1:\mathbf{y}_{j}^{k}=\emptyset}^{n_{\mathbf{Y}}}W^{k}(i,j)\nonumber \\
 & +\rho c^{p}\sum_{j=1:\mathbf{y}_{j}^{k}\neq\emptyset}^{n_{\mathbf{Y}}}W^{k}(n_{\mathbf{X}}+1,j)\nonumber \\
 & +\left(1-\rho\right)c^{p}\sum_{i=1:\mathbf{x}_{i}^{k}\neq\emptyset}^{n_{\mathbf{X}}}W^{k}(i,n_{\mathbf{Y}}+1).
\end{align}

Now, we use (\ref{eq:binary_constraint2}) and (\ref{eq:binary_constraint3})
to remove $W^{k}(n_{\mathbf{X}}+1,j)$ and $W^{k}(i,n_{\mathbf{Y}}+1)$
from the previous equation. By simplifying and rearranging the terms,
we obtain
\begin{align}
\mathrm{tr}\big[\big(D_{\mathbf{X},\mathbf{Y}}^{k}\big)^{\dagger}W^{k}\big] & =\sum_{i=1:\mathbf{x}_{i}^{k}\neq\emptyset}^{n_{\mathbf{X}}}\sum_{j=1:\mathbf{y}_{j}^{k}\neq\emptyset}^{n_{\mathbf{Y}}}W^{k}(i,j)D_{\mathbf{X},\mathbf{Y}}^{k}(i,j)\nonumber \\
 & +\sum_{j=1:\mathbf{y}_{j}^{k}\neq\emptyset}^{n_{\mathbf{Y}}}\rho c^{p}+\sum_{i=1:\mathbf{x}_{i}^{k}\neq\emptyset}^{n_{\mathbf{X}}}\left(1-\rho\right)c^{p}\nonumber \\
 & +\rho c^{p}\sum_{i=1:\mathbf{x}_{i}^{k}=\emptyset}^{n_{\mathbf{X}}}\sum_{j=1:\mathbf{y}_{j}^{k}\neq\emptyset}^{n_{\mathbf{Y}}}W^{k}(i,j)\nonumber \\
 & -\rho c^{p}\sum_{j=1:\mathbf{y}_{j}^{k}\neq\emptyset}^{n_{\mathbf{Y}}}\sum_{i=1}^{n_{\mathbf{X}}}W^{k}(i,j)\nonumber \\
 & +\left(1-\rho\right)c^{p}\sum_{i=1:\mathbf{x}_{i}^{k}\neq\emptyset}^{n_{\mathbf{X}}}\sum_{j=1:\mathbf{y}_{j}^{k}=\emptyset}^{n_{\mathbf{Y}}}W^{k}(i,j)\nonumber \\
 & -\left(1-\rho\right)c^{p}\sum_{i=1:\mathbf{x}_{i}^{k}\neq\emptyset}^{n_{\mathbf{X}}}\sum_{j=1}^{n_{\mathbf{Y}}}W^{k}(i,j).
\end{align}

We can simplify the third and fourth line, and the fifth and sixth
line as follows
\begin{align}
\mathrm{tr}\big[\big(D_{\mathbf{X},\mathbf{Y}}^{k}\big)^{\dagger}W^{k}\big] & =\sum_{i=1:\mathbf{x}_{i}^{k}\neq\emptyset}^{n_{\mathbf{X}}}\sum_{j=1:\mathbf{y}_{j}^{k}\neq\emptyset}^{n_{\mathbf{Y}}}W^{k}(i,j)D_{\mathbf{X},\mathbf{Y}}^{k}(i,j)\nonumber \\
 & +\sum_{j=1:\mathbf{y}_{j}^{k}\neq\emptyset}^{n_{\mathbf{Y}}}\rho c^{p}+\sum_{i=1:\mathbf{x}_{i}^{k}\neq\emptyset}^{n_{\mathbf{X}}}\left(1-\rho\right)c^{p}\nonumber \\
 & -\rho c^{p}\sum_{j=1:\mathbf{y}_{j}^{k}\neq\emptyset}^{n_{\mathbf{Y}}}\sum_{i=1:\mathbf{x}_{i}^{k}\neq\emptyset}^{n_{\mathbf{X}}}W^{k}(i,j)\nonumber \\
 & -\left(1-\rho\right)c^{p}\sum_{i=1:\mathbf{x}_{i}^{k}\neq\emptyset}^{n_{\mathbf{X}}}\sum_{j=1:\mathbf{y}_{j}^{k}\neq\emptyset}^{n_{\mathbf{Y}}}W^{k}(i,j).
\end{align}
Summing the last two lines, we obtain
\begin{align}
\mathrm{tr}\big[\big(D_{\mathbf{X},\mathbf{Y}}^{k}\big)^{\dagger}W^{k}\big] & =\sum_{i=1:\mathbf{x}_{i}^{k}\neq\emptyset}^{n_{\mathbf{X}}}\sum_{j=1:\mathbf{y}_{j}^{k}\neq\emptyset}^{n_{\mathbf{Y}}}W^{k}(i,j)D_{\mathbf{X},\mathbf{Y}}^{k}(i,j)\nonumber \\
 & +\sum_{j=1:\mathbf{y}_{j}^{k}\neq\emptyset}^{n_{\mathbf{Y}}}\rho c^{p}+\sum_{i=1:\mathbf{x}_{i}^{k}\neq\emptyset}^{n_{\mathbf{X}}}\left(1-\rho\right)c^{p}\nonumber \\
 & -c^{p}\sum_{j=1:\mathbf{y}_{j}^{k}\neq\emptyset}^{n_{\mathbf{Y}}}\sum_{i=1:\mathbf{x}_{i}^{k}\neq\emptyset}^{n_{\mathbf{X}}}W^{k}(i,j).\label{eq:T_GOSPA_append1}
\end{align}
We can see that the only terms that depend on $\rho$ are those in
the second line. In the first line $D_{\mathbf{X},\mathbf{Y}}^{k}(i,j)$
does not depend on $\rho$ since objects exist. Plugging (\ref{eq:T_GOSPA_append1})
into (\ref{eq:LP_quasimetric}), the terms that depend on $\rho$
can be taken out of the optimisation function, and the optimisation
function does not depend on $\rho$. This completes the proof of Lemma
\ref{lem:Optimal_assignment_TGOSPA_q}. Note that this proof is valid
for the LP T-GOSPA q-metric and also for the T-GOSPA q-metric (without
LP relaxation).

\subsection{Proof of Lemma \ref{lem:Symmetrisation_TGOSPA}\label{subsec:Appendix_D2}}

In this appendix, we prove Lemma \ref{lem:Symmetrisation_TGOSPA}.
Plugging (\ref{eq:T_GOSPA_append1}) into (\ref{eq:LP_quasimetric}),
we obtain
\begin{align}
 & \overline{d}_{p}^{\left(c,\rho,\gamma\right)}\left(\mathbf{X},\mathbf{Y}\right)\nonumber \\
 & =\sum_{k=1}^{T}\left[\sum_{j=1:\mathbf{y}_{j}^{k}\neq\emptyset}^{n_{\mathbf{Y}}}\rho c^{p}+\sum_{i=1:\mathbf{x}_{i}^{k}\neq\emptyset}^{n_{\mathbf{X}}}\left(1-\rho\right)c^{p}\right]\nonumber \\
 & \quad+\min_{\substack{W^{k}\in\mathcal{\overline{W}}_{\mathbf{X},\mathbf{Y}}\\
k=1,\ldots,T
}
}\Bigg(\sum_{k=1}^{T}\left[\sum_{i=1:\mathbf{x}_{i}^{k}\neq\emptyset}^{n_{\mathbf{X}}}\sum_{j=1:\mathbf{y}_{j}^{k}\neq\emptyset}^{n_{\mathbf{Y}}}W^{k}(i,j)D_{\mathbf{X},\mathbf{Y}}^{k}(i,j)\right.\nonumber \\
 & \quad\left.-c^{p}\sum_{j=1:\mathbf{y}_{j}^{k}\neq\emptyset}^{n_{\mathbf{Y}}}\sum_{i=1:\mathbf{x}_{i}^{k}\neq\emptyset}^{n_{\mathbf{X}}}W^{k}(i,j)\right]\nonumber \\
 & \quad+\frac{\gamma^{p}}{2}\sum_{k=1}^{T-1}\sum_{i=1}^{n_{\mathbf{X}}}\sum_{j=1}^{n_{\mathbf{Y}}}|W^{k}(i,j)-W^{k+1}(i,j)|\Bigg)^{\frac{1}{p}}.
\end{align}
Using Lemma \ref{lem:TGOSPA_q_symmetry}, $\overline{d}_{p}^{\left(c,\rho,\gamma\right)}\left(\mathbf{Y},\mathbf{X}\right)=\overline{d}_{p}^{\left(c,1-\rho,\gamma\right)}\left(\mathbf{X},\mathbf{Y}\right)$
and therefore
\begin{align}
 & \overline{d}_{p}^{\left(c,\rho,\gamma\right)}\left(\mathbf{Y},\mathbf{X}\right)\nonumber \\
 & =\sum_{k=1}^{T}\left[\sum_{j=1:\mathbf{y}_{j}^{k}\neq\emptyset}^{n_{\mathbf{Y}}}\left(1-\rho\right)c^{p}+\sum_{i=1:\mathbf{x}_{i}^{k}\neq\emptyset}^{n_{\mathbf{X}}}\rho c^{p}\right]\nonumber \\
 & \quad+\min_{\substack{W^{k}\in\mathcal{\overline{W}}_{\mathbf{X},\mathbf{Y}}\\
k=1,\ldots,T
}
}\Bigg(\sum_{k=1}^{T}\left[\sum_{i=1:\mathbf{x}_{i}^{k}\neq\emptyset}^{n_{\mathbf{X}}}\sum_{j=1:\mathbf{y}_{j}^{k}\neq\emptyset}^{n_{\mathbf{Y}}}W^{k}(i,j)D_{\mathbf{X},\mathbf{Y}}^{k}(i,j)\right.\nonumber \\
 & \quad\left.-c^{p}\sum_{j=1:\mathbf{y}_{j}^{k}\neq\emptyset}^{n_{\mathbf{Y}}}\sum_{i=1:\mathbf{x}_{i}^{k}\neq\emptyset}^{n_{\mathbf{X}}}W^{k}(i,j)\right]\nonumber \\
 & \quad+\frac{\gamma^{p}}{2}\sum_{k=1}^{T-1}\sum_{i=1}^{n_{\mathbf{X}}}\sum_{j=1}^{n_{\mathbf{Y}}}|W^{k}(i,j)-W^{k+1}(i,j)|\Bigg)^{\frac{1}{p}}.
\end{align}

Substituting the above two equations in (\ref{eq:symmetrisation_TGOSPAq_2}),
we obtain the (LP) T-GOSPA metric $\overline{d}_{p}^{\left(c,1/2,\gamma\right)}\left(\mathbf{X},\mathbf{Y}\right)$.
The same result can be obtained equivalently for $d_{p}^{\left(c,1/2,\gamma\right)}\left(\mathbf{X},\mathbf{Y}\right)$
proving Lemma \ref{lem:Symmetrisation_TGOSPA}.

\section{\label{sec:Proof_metric_preserving_mapping}}

In this appendix, for completeness, we prove Lemma \ref{lem:Bounding_metric}
following \cite{Corazza99}. First, we can see that $f\left(d\left(\cdot,\cdot\right)\right)$
is non-negative since $f\left(\cdot\right)$ is non-negative. The
identity property holds directly due to P1. Note that it is not enough
that $f(0)=0$ to meet the identity property. 

The triangle inequality proof is as follows. As $d\left(\cdot,\cdot\right)$
is a metric
\begin{align}
d\left(\mathbf{X},\mathbf{Y}\right) & \leq d\left(\mathbf{X},\mathbf{Z}\right)+d\left(\mathbf{Z},\mathbf{X}\right).
\end{align}
Due to P2, it holds that
\begin{align}
f\left(d\left(\mathbf{X},\mathbf{Y}\right)\right) & \leq f\left(d\left(\mathbf{X},\mathbf{Z}\right)+d\left(\mathbf{Z},\mathbf{X}\right)\right).
\end{align}
Due to P3, we obtain

\begin{align}
f\left(d\left(\mathbf{X},\mathbf{Y}\right)\right) & \leq f\left(d\left(\mathbf{X},\mathbf{Z}\right)\right)+f\left(d\left(\mathbf{Z},\mathbf{X}\right)\right).
\end{align}
This completes the proof of the triangle inequality for $f\left(d\left(\cdot,\cdot\right)\right)$.
Finally, it is also direct from P1, P2 and P4 that the mapping takes
values in $\left[0,1\right]$. 
\end{document}